\pdfoutput=1
\documentclass[11pt]{article}

\usepackage[final]{acl}

\usepackage{times}
\usepackage{latexsym}
\usepackage[dvipsnames]{xcolor}
\usepackage[table]{xcolor}
\usepackage[T1]{fontenc}
\usepackage[utf8]{inputenc}

\usepackage{microtype}

\usepackage{inconsolata}

\usepackage{graphicx}
\usepackage{verbatim}
\usepackage{amsmath}
\usepackage{multirow}
\usepackage{multicol}
\usepackage{booktabs}
\usepackage{enumitem}
\usepackage{array}
\usepackage{xcolor}
\usepackage{algorithm}       % For algorithm environment  
\usepackage{algorithmic}
\usepackage{amssymb}
\usepackage{tabularx}
\usepackage{threeparttable}
\usepackage{subcaption}  % Add this to your preamble
\usepackage{colortbl}
\usepackage{tikz}
\usepackage{xcolor,colortbl}  % for the highlighted version
\definecolor{removecolor}{RGB}{255,200,200} % light red
\definecolor{addcolor}{RGB}{200,255,200}    % light green

\newcommand{\removed}[1]{\colorbox{removecolor}{\textit{#1}}}
\newcommand{\added}[1]{\colorbox{addcolor}{\textit{#1}}}
\definecolor{np1}{HTML}{F4A261} % Warm Apricot
\definecolor{np2}{HTML}{2A9D8F} % Muted Teal
\definecolor{np3}{HTML}{90A955} % Sage Green
\definecolor{np4}{HTML}{A084CA} % Dusty Lavender
\definecolor{np5}{HTML}{5E81AC} % Slate Blue
\definecolor{np6}{HTML}{D9A441} % Soft Ochre
\definecolor{np7}{HTML}{A97155} % Cocoa Mocha

\definecolor{dogcolor}{RGB}{173,216,230}    % light blue
\definecolor{mailmancolor}{RGB}{255,218,185} % light peach

\newcommand{\dog}[1]{\colorbox{dogcolor}{\textbf{#1}}}
\newcommand{\mailman}[1]{\colorbox{mailmancolor}{\textbf{#1}}}
\definecolor{highlight}{RGB}{200, 230, 201}  % Light green for highest values
\newcommand{\np}[2]{\textbf{\textcolor{#1}{#2}}}
\newcommand{\npmark}[2]{\(\mathrm{#1}_{#2}\)}

\usepackage{tcolorbox}
\usepackage{xcolor}
\usepackage{soul} % for highlighting

\definecolor{questionblue}{RGB}{173,216,230}
\definecolor{answerblue}{RGB}{173,216,230} 
\definecolor{thinkingpink}{RGB}{255,192,203}
\definecolor{thinkinggreen}{RGB}{200,255,200} 
\definecolor{highlightpink}{RGB}{255,105,180}
\definecolor{highlightred}{RGB}{255,200,200}
\definecolor{highlightyellow}{RGB}{255,255,180} 

\sethlcolor{highlightpink}

\newcommand{\hlred}[1]{\sethlcolor{highlightred}\hl{#1}}
\newcommand{\hlyellow}[1]{\sethlcolor{highlightyellow}\hl{#1}}
\newcommand{\hlgreen}[1]{\sethlcolor{thinkinggreen}\hl{#1}}
\newcommand{\hlblue}[1]{\sethlcolor{answerblue}\hl{#1}}

\newcommand{\tableline}[1]{\par\vspace{0.5ex}#1}

\newcommand{\corpusname}{\textsc{CenterBench}}
\title{The Dog the Cat Chased Stumped the Model: \\ Measuring When Language Models Abandon Structure for Shortcuts}

\author{Sangmitra Madhusudan$^1$\textnormal{,} Kaige Chen$^2$\textnormal{,} \and Ali Emami$^1$ \\
  $^1$Emory University, Atlanta, USA \\
  $^2$Brock University, St. Catharines, Canada \\
  \texttt{smadhus@emory.edu, ax23ev@brocku.ca, aemami@emory.edu} \\}
  
\begin{document}
\maketitle
\begin{abstract}
When language models correctly parse ``The cat that the dog chased meowed,'' are they analyzing syntax or simply familiar with dogs chasing cats? Despite extensive benchmarking, we lack methods to distinguish structural understanding from semantic pattern matching. We introduce \textbf{\corpusname{}}, a dataset of 9,720 comprehension questions on center-embedded sentences (like ``The cat [that the dog chased] meowed'') where relative clauses nest recursively, creating processing demands from simple to deeply nested structures. Each sentence has a syntactically identical but semantically implausible counterpart (e.g., mailmen prescribe medicine, doctors deliver mail) and six comprehension questions testing surface understanding, syntactic dependencies, and causal reasoning. Testing six models reveals that performance gaps between plausible and implausible sentences widen systematically with complexity, with models showing median gaps up to 26.8 percentage points, quantifying when they abandon structural analysis for semantic associations. Notably, semantic plausibility harms performance on questions about resulting actions, where following causal relationships matters more than semantic coherence. Reasoning models improve accuracy but their traces show semantic shortcuts, overthinking, and answer refusal. Unlike models whose plausibility advantage systematically widens with complexity, humans shows variable semantic effects. \corpusname{} provides the first framework to identify when models shift from structural analysis to pattern matching.\footnote{The complete dataset and codebase are publicly accessible on \href{https://github.com/Sangmitra-06/CENTERBENCH}{GitHub}.}
\end{abstract}

\section{Introduction}

Large language models (LLMs) can explain quantum mechanics and write sophisticated code, yet often fail to parse sentences like ``The cat that the dog that the mouse feared chased meowed'', a construction that any undergraduate in linguistics can successfully diagram. This striking contrast points to a fundamental uncertainty in how we evaluate model capabilities: When models produce correct answers, are they genuinely parsing syntactic structure or merely exploiting semantic associations? %As we increasingly deploy these systems in high-stakes applications, distinguishing true linguistic understanding from pattern matching becomes critical for assessing their reliability.

\begin{figure}[t]
    \centering
    \includegraphics[width=1.0\linewidth]{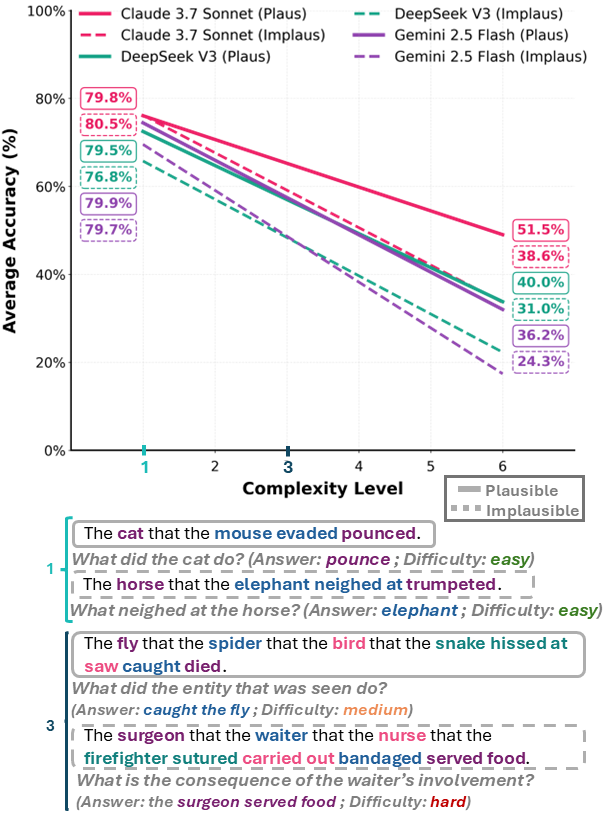}
   
    \caption{Performance degradation across complexity levels for three models on plausible vs.\ implausible center-embedded sentences (averaged across all question types), with example questions of varying difficulty.}
    \label{fig:main-figure}
\end{figure}

Standard benchmarks like MMLU \cite{hendrycks2021measuringmassivemultitasklanguage} and HLE \cite{phan2025humanitysexam} cannot address this uncertainty—they measure only final accuracy, revealing nothing on whether models arrived at correct answers through structural analysis or semantic pattern matching. We need evaluation frameworks that can (1) distinguish structural understanding from semantic shortcuts and (2) track how this balance shifts with increasing complexity.

%Yet distinguishing genuine understanding from pattern matching proves surprisingly difficult. Traditional benchmarks like MMLU \cite{hendrycks2021measuringmassivemultitasklanguage} and GPQA \cite{rein2023gpqagraduatelevelgoogleproofqa} measure only final accuracy, revealing nothing about the reasoning process. When we consider how reasoning might degrade with increasing complexity, the challenge grows. Recent work by \citet{shojaee2025illusionthinkingunderstandingstrengths} claimed models experience ``reasoning collapse'' on increasingly complex planning puzzles, but \citet{lawsen2025commentillusionthinkingunderstanding} showed these failures were evaluation artifacts: models succeeded when asked to generate algorithms rather than exhaustive move lists, and some puzzles were, in fact, unsolvable. Beyond methodological issues, this highlights our core problem: when models \textit{do} succeed on language tasks, we cannot tell whether they truly understand structure or merely exploit patterns. We need a framework that \textbf{(1) distinguishes structural understanding from semantic shortcuts} and \textbf{(2) tracks how this balance shifts with complexity}.

%Center-embedded sentences, where relative clauses nest recursively within each other, offer an ideal solution \cite{Chomsky1963-CHOITT-2}. Consider how complexity builds naturally:

\begin{table*}[!t]
\centering
\small
\resizebox{0.9\textwidth}{!}{%
\renewcommand{\arraystretch}{1.3}
\begin{tabular}{@{}p{0.7cm}p{1.4cm}p{17cm}@{}}
\toprule
\textbf{Level} & \textbf{Subset} & \textbf{Sentences} \\
\midrule

1 & Plausible & 
The \np{np1}{cat} that the \np{np2}{mouse} \np{np2}{evaded} \np{np1}{pounced}. \\
& & \textit{
\npmark{NP}{1}: cat,\quad
\npmark{NP}{2}: mouse,\quad
\npmark{V}{2}: evaded,\quad
\npmark{V}{1}: pounced} \\

\addlinespace[1ex]

1 & Implausible & 
The \np{np1}{horse} that the \np{np2}{elephant} \np{np2}{neighed at} \np{np1}{trumpeted}. \\
& & \textit{
\npmark{NP}{1}: horse,\quad
\npmark{NP}{2}: elephant,\quad
\npmark{V}{2}: neighed at,\quad
\npmark{V}{1}: trumpeted} \\

\addlinespace[1.5ex]

2 & Plausible & 
The \np{np1}{bicycle} that the \np{np2}{car} that the \np{np3}{truck} \np{np3}{hit} \np{np2}{bumped} \np{np1}{fell over}. \\
& & \textit{
\npmark{NP}{1}: bicycle,\quad
\npmark{NP}{2}: car,\quad
\npmark{NP}{3}: truck,\quad
\npmark{V}{3}: hit,\quad
\npmark{V}{2}: bumped,\quad
\npmark{V}{1}: fell over} \\

\addlinespace[1ex]

2 & Implausible & 
The \np{np1}{bicycle} that the \np{np2}{airplane} that the \np{np3}{train} \np{np3}{pedaled past} \np{np2}{whistled at} \np{np1}{taxied}. \\
& & \textit{
\npmark{NP}{1}: bicycle,\quad
\npmark{NP}{2}: airplane,\quad
\npmark{NP}{3}: train,\quad
\npmark{V}{3}: pedaled past,\quad
\npmark{V}{2}: whistled at,\quad
\npmark{V}{1}: taxied} \\

\addlinespace[1.5ex]

3 & Plausible & 
The \np{np1}{fly} that the \np{np2}{spider} that the \np{np3}{bird} that the \np{np4}{snake} \np{np4}{hissed at} \np{np3}{saw} \np{np2}{caught} \np{np1}{died}. \\
& & \textit{
\npmark{NP}{1}: fly,\quad
\npmark{NP}{2}: spider,\quad
\npmark{NP}{3}: bird,\quad
\npmark{NP}{4}: snake,\quad
\npmark{V}{4}: hissed at,\quad
\npmark{V}{3}: saw,\quad
\npmark{V}{2}: caught,\quad
\npmark{V}{1}: died} \\

\addlinespace[1ex]

3 & Implausible & 
The \np{np1}{skunk} that the \np{np2}{lion} that the \np{np3}{bat} that the \np{np4}{elephant} \np{np4}{sprayed at} \np{np3}{trumpeted at} \np{np2}{echolocated} \np{np1}{roared}. \\
& & \textit{
\npmark{NP}{1}: skunk,\quad
\npmark{NP}{2}: lion,\quad
\npmark{NP}{3}: bat,\quad
\npmark{NP}{4}: elephant,\quad
\npmark{V}{4}: sprayed at,\quad
\npmark{V}{3}: trumpeted at,\quad
\npmark{V}{2}: echolocated,\quad
\npmark{V}{1}: roared} \\

\bottomrule
\end{tabular}}

\caption{Example sentences from complexity levels 1-3, with color-coded noun-verb pairs for plausible and implausible subsets. Examples for complexity levels 1-6 are provided in Appendix Table~\ref{appendix:sentence-complexity1-6}.}
\label{tab:sentence-complexity1-3}
\end{table*}

Center-embedded sentences \cite{Chomsky1963-CHOITT-2} offer an ideal testbed. In these constructions, relative clauses nest recursively, creating natural complexity scaling:

\begin{quote}
\small

\textbf{Level 1:} The \textcolor{blue}{cat} [that the \textcolor{red}{dog} \textcolor{red}{chased}] \textcolor{blue}{meowed}.\\
\textbf{Level 2:} The \textcolor{blue}{cat} [that the \textcolor{red}{dog} [that the \textcolor{ForestGreen}{boy} \textcolor{ForestGreen}{saw}] \textcolor{red}{chased}] \textcolor{blue}{meowed}.\\
\textbf{Level 3:} The \textcolor{blue}{cat} [that the \textcolor{red}{dog} [that the \textcolor{ForestGreen}{boy} [that the \textcolor{purple}{girl} \textcolor{purple}{liked}] \textcolor{ForestGreen}{saw}] \textcolor{red}{chased}] \textcolor{blue}{meowed}.

\end{quote}

Unlike artificial puzzles, these constructions appear in natural language, have well-defined syntactic parses, and show consistent human processing patterns established through decades of psycholinguistic research \cite{Miller1963-MILFMO, Gibson1998-xu}. Deep embeddings rarely appear in training corpora \cite{Karlsson2007}, minimizing memorization. While \citet{hardt2025sparks} found GPT-4 maintains accuracy on these structures up to 3-4 levels, they didn't investigate whether this stemmed from structural analysis or semantic strategies.

To address this gap, we create matched sentence pairs differing only in semantic plausibility:

\begin{quote}

\hypertarget{1a}{\textbf{(1a)} Plausible:} The \textcolor{blue}{cat} that the \textcolor{red}{dog} \textcolor{red}{chased} \textcolor{blue}{meowed}.\\
\hypertarget{1b}{\textbf{(1b)} Implausible:} The \textcolor{blue}{waiter} that the \textcolor{red}{mailman} \textcolor{red}{seated} \textcolor{blue}{delivered mail}.

\end{quote}

Both sentences share identical syntax: parsing who did what requires tracking the same color-coded noun-verb relationships. The crucial difference lies in semantic plausibility. While real-world knowledge supports dogs chasing cats, it contradicts mailmen seating waiters. By comparing performance across matched pairs at increasing complexity levels (1-6), we can quantify exactly when and how models transition from structural analysis to semantic shortcuts. This extends \citet{wilcox-etal-2019-hierarchical}'s single-depth plausible/implausible design to systematic complexity scaling.

We introduce \textbf{\corpusname{}}, a benchmark of 360 center-embedded sentences with 9,720 comprehension questions, featuring controlled complexity scaling and plausible/implausible pairing. Evaluating six models reveals notable patterns: performance degrades consistently with complexity (\S\ref{result-1}), while the plausible-implausible gap widens systematically, reaching median values over 25 percentage points (\S\ref{result-2}, Figure~\ref{fig:main-figure}). Reasoning demands also matter: tasks involving causal reasoning show reversed patterns where semantics hinder rather than help (\S\ref{result-3}). Reasoning models improve accuracy but their traces suggest persistent failures: choosing semantic plausibility over syntactic structure, refusing to report implausible relationships, and overthinking (\S\ref{result-4} 
\& \S\ref{result-5}). Human comparisons reveal inconsistent semantic effects, with plausibility helping at some complexity levels but hindering at others (\S\ref{result-6}).

%These findings have immediate practical implications for deploying language models in applications requiring genuine structural understanding. By providing the first framework to identify precisely when models abandon structural analysis for shortcuts, \corpusname{} enables more informed decisions about model deployment in critical domains.

%By mapping when and how models transition from structural analysis to shortcuts, \corpusname{} provides critical insights into language understanding limits. Our findings suggest current models lack robust mechanisms for maintaining syntactic parsing when semantic cues become unreliable, with serious implications for applications requiring genuine language understanding.

%This work makes four key contributions:

%\begin{enumerate}
%\item \textbf{\corpusname{}}: A benchmark of 9,720 center-embedded sentences with controlled complexity and systematic plausible/implausible pairing, enabling precise measurement of semantic shortcut reliance

%\item \textbf{Empirical evidence} that models increasingly abandon syntactic parsing for semantic shortcuts as complexity grows, with the performance gap directly quantifying this transition (Figure~\ref{fig:main-figure})

%\item \textbf{Analysis of reasoning failures} through Chain-of-Thought traces, revealing circular logic, hallucinated structures, and overconfident errors

%\item \textbf{Human-model comparisons} demonstrating fundamental differences in processing strategies under complexity
%\end{enumerate}

\section{\corpusname}
\label{sec:dataset}

\corpusname{} consists of 9,720 questions across 360 center-embedded sentences designed to test whether language models truly understand syntactic structure or rely on semantic shortcuts.

As illustrated in examples \hyperlink{1a}{\textbf{(1a)}}
and \hyperlink{1b}{\textbf{(1b)}}, we create sentence pairs with identical syntax but different semantic plausibility, then ask questions that require tracking noun-verb relationships. For both sentences, asking ``What did the \textcolor{blue}{cat}/\textcolor{blue}{waiter} do?'' requires linking the first noun with the last verb (\textcolor{blue}{``meowed''}, \textcolor{blue}{``delivered mail''}). Similarly, ``Who \textcolor{red}{chased}/\textcolor{red}{seated} the \textcolor{blue}{cat}/\textcolor{blue}{waiter}?'' tests whether models correctly identify \textcolor{red}{``the dog''}/\textcolor{red}{``the mailman''}. If models perform worse on questions involving (1b) despite identical syntactic structure to (1a), they must be using semantic shortcuts rather than structural understanding.

\corpusname{} scales this approach across complexity levels 1-6, where each level adds one nested relative clause. Table \ref{tab:sentence-complexity1-3} shows how structural complexity increases while maintaining the plausible/implausible contrast. Each sentence includes six comprehension questions per entity, ranging from simple action identification to complex causal reasoning. The dataset has 4,860 questions each for plausible and implausible conditions (9,720 total), with composition details in Table \ref{tab:dataset_composition}.

\subsection{Sentence Creation}

Each sentence in \corpusname\ follows the following structure:

\[
\begin{aligned}
S(n) =\ & NP_1\; [\text{that } NP_2\; [\dots\; [\text{that } NP_{n+1}\; V_{n+1}] \\
       & \quad \dots\, V_2]\, V_1
\end{aligned}
\]

Here, $S(n)$ represents a sentence at the complexity level $n$. Each $NP_i$ stands for a \textit{noun phrase}, such as ``the cat'' or ``the dog'' and each $V_i$ stands for a \textit{verb}, such as ``chased'' or ``barked''. The final verb, $V_1$, is always \textit{intransitive}, while all intermediate verbs ($V_2, V_3, \ldots, V_{n+1}$) are \textit{transitive} in nature. This creates the parsing challenge: Models must track which noun performs which action across the nested structure.

\subsubsection{Plausible Subset}

\label{sec: sentence-plausible}
For plausible sentences, we selected nouns from established psycholinguistic norms \cite{VANOVERSCHELDE2004289} in three categories: animals, people (occupations) and vehicles. We refined these lists to include only common, everyday nouns, for example, keeping ``eagle'' and ``crow'' while excluding ``oriole'' and ``chickadee''. %(see Appendix Table \ref{appendix:noun-inclusion} for complete lists).

We generated 30 sentences per complexity level using \textit{GPT-4-0613} in a one-shot setting, providing our curated noun lists as a foundation while allowing flexibility for natural sentence construction. The model occasionally incorporated related nouns (e.g. ``hawk'' or ``bird'') to create more fluent sentences. The complete noun inventory and all generation prompts appear in Appendix Sections \ref{sec: nouns_and_verbs} and \ref{sec: sentence_generation}, respectively.

%, allowing it to incorporate additional nouns when constructing sentences.

\subsubsection{Implausible Subset}

\label{sec: sentence-implausible}

To create semantically implausible sentences with identical syntactic structure, we used a systematic verb-swapping approach. Starting with our plausible noun list, we refined it to include only entities with maximally distinct behaviors. For example, from people-nouns, we retained ``doctor,'' ``police officer,'' and ``lawyer'' while removing ``superintendent'' and ``veterinarian'' whose actions overlap with other professions.
\begin{table}[!t]
\centering
\resizebox{0.85\columnwidth}{!}{%
\begin{tabular}{ccccc}
\hline
\textbf{Complexity} & \textbf{Easy} & \textbf{Medium} & \textbf{Hard} & \textbf{Total} \\
\hline
1 & 120 & 120 & 120 & 360 \\
2 & 180 & 180 & 180 & 540 \\
3 & 240 & 240 & 240 & 720 \\
4 & 300 & 300 & 300 & 900 \\
5 & 360 & 360 & 360 & 1080 \\
6 & 420 & 420 & 420 & 1260 \\
\hline
\multicolumn{4}{c}{\textbf{Subtotal per subset}} & \textbf{4860} \\
\multicolumn{4}{c}{\textbf{Total dataset (plausible + implausible)}} & \textbf{9720} \\
\hline
\end{tabular}%
}

\caption{\corpusname{} Composition}
\label{tab:dataset_composition}
\end{table}

We then used \textit{claude-sonnet-4-20250514} through the Anthropic Console\footnote{\url{https://console.anthropic.com/workbench}} to generate unique verbs for each retained entity. For example:
\begin{itemize}[itemsep=0pt, parsep=0pt, topsep=0pt, partopsep=0pt,leftmargin=*]
    \item \textbf{doctor}: \textit{prescribed medicine to} (transitive), \textit{diagnosed} (intransitive)
    \item \textbf{police officer}: \textit{arrested} (transitive), \textit{patrolled} (intransitive)
    \item \textbf{lawyer}: \textit{cross-examined} (transitive), \textit{objected} (intransitive)
\end{itemize}

\begin{table*}[!t]
\centering
\small
\renewcommand{\arraystretch}{0.5}
\setlength{\tabcolsep}{6pt}
\resizebox{0.85\textwidth}{!}{
\begin{tabular}{@{}>{\raggedright\arraybackslash}p{1.6cm} >{\raggedright\arraybackslash}p{3.2cm} >{\raggedright\arraybackslash}p{5.5cm} >{\raggedright\arraybackslash}p{5.5cm}@{}}
\toprule
\multicolumn{4}{c}{\textbf{Example Sentence:} The \dog{dog} that the \mailman{mailman} startled barked.} \\
\midrule
\textbf{Category} & \textbf{Question Type} & \textbf{Questions about \dog{dog}} & \textbf{Questions about \mailman{mailman}} \\
\midrule
\multirow{2}{*}{\textbf{Easy}} 
& Action Performed & What did the \dog{dog} do? → bark & What did the \mailman{mailman} do? → startle the dog \\
& Agent Identification & Who startled the \dog{dog}? → the mailman & What was affected by the \mailman{mailman}? → the dog \\
\midrule
\addlinespace
\multirow{2}{*}{\textbf{Medium}} 
& Entity Count & How many distinct entities are in the sentence? → 2 & How many distinct entities are in the sentence? → 2 \\
& Nested Dependency & What did the entity that was startled do? → barked & What did the entity acted upon by the \mailman{mailman} do? → barked \\
\midrule
\addlinespace
\multirow{2}{*}{\textbf{Hard}} 
& Causal Sequence & What series of events led to the \dog{dog}'s action? → the mailman startling the dog & What series of events led to the \mailman{mailman}'s action? → no prior events \\
& Chain Consequence & What is the consequence of the \dog{dog}'s involvement? → none & What is the consequence of the \mailman{mailman}'s involvement? → the dog barked \\
\bottomrule
\end{tabular}}

\caption{Question difficulty levels and types illustrated with entity-specific questions for a complexity level 1 center-embedded sentence. Color coding distinguishes the two entities: \dog{dog} and \mailman{mailman}.}
\label{tab:example-questions}
\end{table*}

To generate implausible sentences, we implemented circular verb swapping: each noun receives verbs originally associated with the next noun in the sequence, with the last noun getting verbs from the first. This ensures semantic violations while maintaining grammatical correctness. For example, in a sentence with [doctor, police officer, lawyer], the doctor would ``patrol,'' the police officer would ``cross-examine,'' and the lawyer would ``diagnose.''

We generated 30 sentences per complexity level using this algorithm, ensuring no duplicate combinations across the dataset. Full algorithmic details and sample verb assignments appear in Appendix Section \ref{sec:algorithm} and Appendix Table \ref{app:entity_verb_samples}, respectively.

\subsection{Sentence validation}

All generated sentences from both subsets underwent manual validation across three dimensions:

\begin{itemize}[itemsep=0pt, parsep=0pt, topsep=0pt, partopsep=0pt, leftmargin=*]
    \item \textbf{Temporal validity}: Entities made inactive (e.g., caught, killed) cannot perform subsequent actions
    \item \textbf{Semantic requirements}: Plausible sentences must contain realistic actions; implausible sentences must violate semantic expectations
    \item \textbf{Syntactic accuracy}: Each sentence must have the correct number of entities and verbs for its complexity level
\end{itemize}

Appendix Table \ref{app:validation-singlecol} illustrates our validation process with examples from complexity level 2. When sentences failed any criterion, we manually corrected them while preserving structure and complexity. For instance, ``caught'' preceding ``chased'' creates a temporal violation (a caught entity cannot chase), which we fixed by substituting ``saw.'' Similarly, ``adopted'' in the plausible subset might accidentally create an implausible relationship, requiring replacement with a proper plausible verb.

Table \ref{tab:sentence-complexity1-3} shows instances from both subsets across complexity levels 1-3, with complete examples through level 6 in Appendix Table \ref{appendix:sentence-complexity1-6}.

\subsection{Question and Answer Generation}

We developed an automated system to generate six comprehension questions and their corresponding answers for each entity in every sentence, grouped into three difficulty levels:

\begin{itemize}[itemsep=0pt, parsep=0pt, topsep=0pt, partopsep=0pt, leftmargin=*]
    \item \textbf{Easy} questions test basic subject-verb relationships (e.g., ``What did the dog do?'')
    \item \textbf{Medium} questions require understanding syntactic structure (e.g., ``What did the entity that was chased do?'')
    \item \textbf{Hard} questions demand forward and backward causal reasoning (e.g., ``What series of events led to the dog's action?'')
\end{itemize}

Table \ref{tab:example-questions} illustrates all the questions with their answers on a single instance. Each question-answer pair targets specific aspects of comprehension, from simple action identification to complex dependency resolution.

Our generation algorithm operates in four steps:
\begin{enumerate}[itemsep=0pt, parsep=0pt, topsep=0pt, partopsep=0pt]
    \item \textbf{Noun identification}: Extracts entities using predefined lists, handling multi-word nouns (e.g., ``police officer'')
    \item \textbf{Structural parsing}: Maps subject-verb-object relationships by analyzing the reversed verb order
    \item \textbf{Verb processing}: Converts verbs to appropriate forms (base, participle, gerund) using morphological rules
    \item \textbf{Template instantiation}: Fills question-answer templates based on parsed relationships, ensuring answers use exact wording from the source sentence
\end{enumerate}

All generated questions and answers underwent manual review. We corrected two primary error types: phrasal verb separation (e.g., ``run away'' split incorrectly) and irregular verb morphology (e.g., ``called'' → ``cal'' instead of ``call''). Complete question-answer templates and a detailed walkthrough are provided in Appendix Section \ref{sec:question_generation}.

\begin{figure}[!t]
    \centering
    \includegraphics[width=1.0\columnwidth]{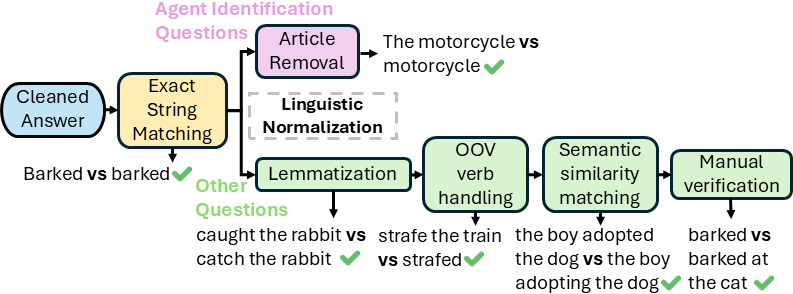}
    
    \caption{Overview of the evaluation pipeline for processing  model responses  (shown before `vs') against gold standard answers}
    \label{fig:flowchart}
\end{figure}

\section{Evaluation Pipeline}

\label{sec: eval}
Our evaluation pipeline assesses whether model responses match gold standard answers through a multi-tier matching strategy. Figure \ref{fig:flowchart} provides an overview of the complete process.

Before evaluation, we preprocess all model responses by removing hidden Unicode characters and stripping common prefixes (e.g., \verb|Answer:|, \verb|**Answer**:|) to ensure consistent formatting across different model outputs.

\subsection{Automatic Evaluation}

We apply increasingly sophisticated matching strategies, proceeding to the next step only if the previous one fails to find a match:

\paragraph{Exact string matching:} Compare lowercase model and gold answers directly. This catches most correct responses immediately.

\paragraph{Linguistic normalization:} For non-matching responses, we apply context-appropriate processing:
\begin{itemize}[itemsep=0pt, parsep=0pt, topsep=0pt, partopsep=0pt,leftmargin=*]
    \item \textit{Agent identification questions}: Remove leading articles (``the,'' ``a,'' ``an'') since ``the mailman'' and ``mailman'' are equivalent answers. If no match is found, the response is marked incorrect.
    \item \textit{Other question types}: Apply lemmatization using spaCy's \verb|en_core_web_sm| model \cite{honnibal2020spacy} to handle verb inflections (e.g., ``chased'' → ``chase''). If no match is found, processing continues to the next step.
\end{itemize}

\paragraph{Out-of-vocabulary (OOV) verb handling:} Several verbs in our dataset (e.g., ``gavel,'' ``neigh'') are missing from spaCy's vocabulary, causing incorrect rejections. We maintain a lookup table of these verbs with their inflected forms (see Appendix Table \ref{tab:missing-verbs}) to ensure correct evaluation. If a match is found here, the response is marked correct.

\paragraph{Semantic similarity matching:} As a final step for remaining unmatched responses, we compute cosine similarity between sentence embeddings using \verb|all-MiniLM-L6-v2| \cite{wang2020minilmdeepselfattentiondistillation}. For example, the gold answer ``the truck hitting the car which led to the car bumping the bicycle'' and model response ``the truck hit the car, the car bumped the bicycle'' convey identical meaning despite different syntax. Responses with similarity $\geq 0.9$ are marked correct, following established QA evaluation practices \cite{morris-etal-2020-reevaluating, berger-etal-2021-dont, six2025decompositionalreasoninggraphretrieval}.\footnote{This step improved evaluation accuracy from $\sim$94\% to 100\% on Claude's responses to 3,240 hard questions.}

\subsection{Manual Verification}

To validate our evaluation pipeline, we manually reviewed the automatically scored responses for the entire plausible subset. Our automatic pipeline achieved 98.95\% scoring accuracy—meaning it agreed with human judgment 98.95\% of the time.

The remaining 1.05\% were evaluation errors (not model errors), concentrated in action performed questions where models provided correct but incomplete answers. For instance, given ``What did the doctor do?'' with gold answer ``cross-examine the lawyer,'' a model response of ``cross-examined'' was incorrectly marked wrong by our system despite being substantially correct. We manually identified and corrected these evaluation errors to ensure accurate model assessment.\footnote{Fleiss's Kappa of 1.0 between two independent annotators}

\begin{comment}
    \begin{table*}[ht]
\centering
\begin{tabularx}{\textwidth}{XXXXX}
\toprule
\textbf{sentence} & \textbf{question} & \textbf{gold answer} & \textbf{model prediction} \\
\midrule
The bicycle that the car that the truck hit bumped fell over. & What series of events led to the bicycle's action? & the truck \textbf{\textit{hitting}} the car \textbf{\textit{which led to}} the car \textbf{\textit{bumping}} the bicycle & the truck \textbf{\textit{hit}} the car, the car \textbf{\textit{bumped}} the bicycle \\
\addlinespace
The pig that the dog that the farmer commanded herded squealed. & What series of events led to the pig's action? & the farmer \textbf{\textit{commanding}} the dog \textbf{\textit{which led to}} the dog \textbf{\textit{herding}} the pig. & the farmer \textbf{\textit{commanded}} the dog the dog \textbf{\textit{herded}} the pig \\
\bottomrule
\end{tabularx}
\caption{Sample valid model predictions evaluated as false due to semantic fallbacks without enabling cosine similarity matching. The questions sampled are causal\_sequence questions for complexity level 2 sentences.}
\label{tab:semantic-fallbacks}
\end{table*}
\end{comment}

\section{Experiments}

\paragraph{Models:} We evaluated three models across different families and capabilities: DeepSeek-V3 \cite{deepseekai2025deepseekv3technicalreport}, Claude 3.7 Sonnet \cite{anthropic-claude3.7}, and Gemini 2.5 Flash \cite{Kavukcuoglu_2025}. For each model, we tested both standard inference and reasoning-enhanced modes: DeepSeek-R1 \cite{deepseekai2025deepseekr1incentivizingreasoningcapability}, Claude 3.7 Sonnet with extended thinking enabled, and Gemini 2.5 Flash with thinking and dynamic thinking activated. Detailed configurations and prompts appear in Appendix Section \ref{sec:model_prompts}.

\paragraph{Human Evaluation:} We conducted a pilot human evaluation using one randomly selected sentence per complexity level (1-4) from each subset (plausible and implausible), totaling 8 sentences. For each sentence, we evaluated all 6 questions targeting one specific entity, using 3 different participants per sentence. To prevent familiarization with the center-embedding format, no participant saw more than one sentence. This resulted in 24 total volunteer participants (3 participants × 8 sentences). Participants completed the study\footnote{\url{https://center-embedding-study.vercel.app/}}, with instructions provided on the form's homepage. %These experiments establish a baseline protocol for future large-scale studies.

\paragraph{Metrics:} We calculated accuracy as the percentage of correct answers using the evaluation pipeline from Section \ref{sec: eval}. Non-thinking models were run 10 times per question and averaged; thinking models were run once due to computational cost. Human responses were evaluated manually.

\section{Results}

\subsection{Performance degrades with complexity}
\label{result-1}

\begin{figure}[!t]
    \centering
    \includegraphics[width=\linewidth]{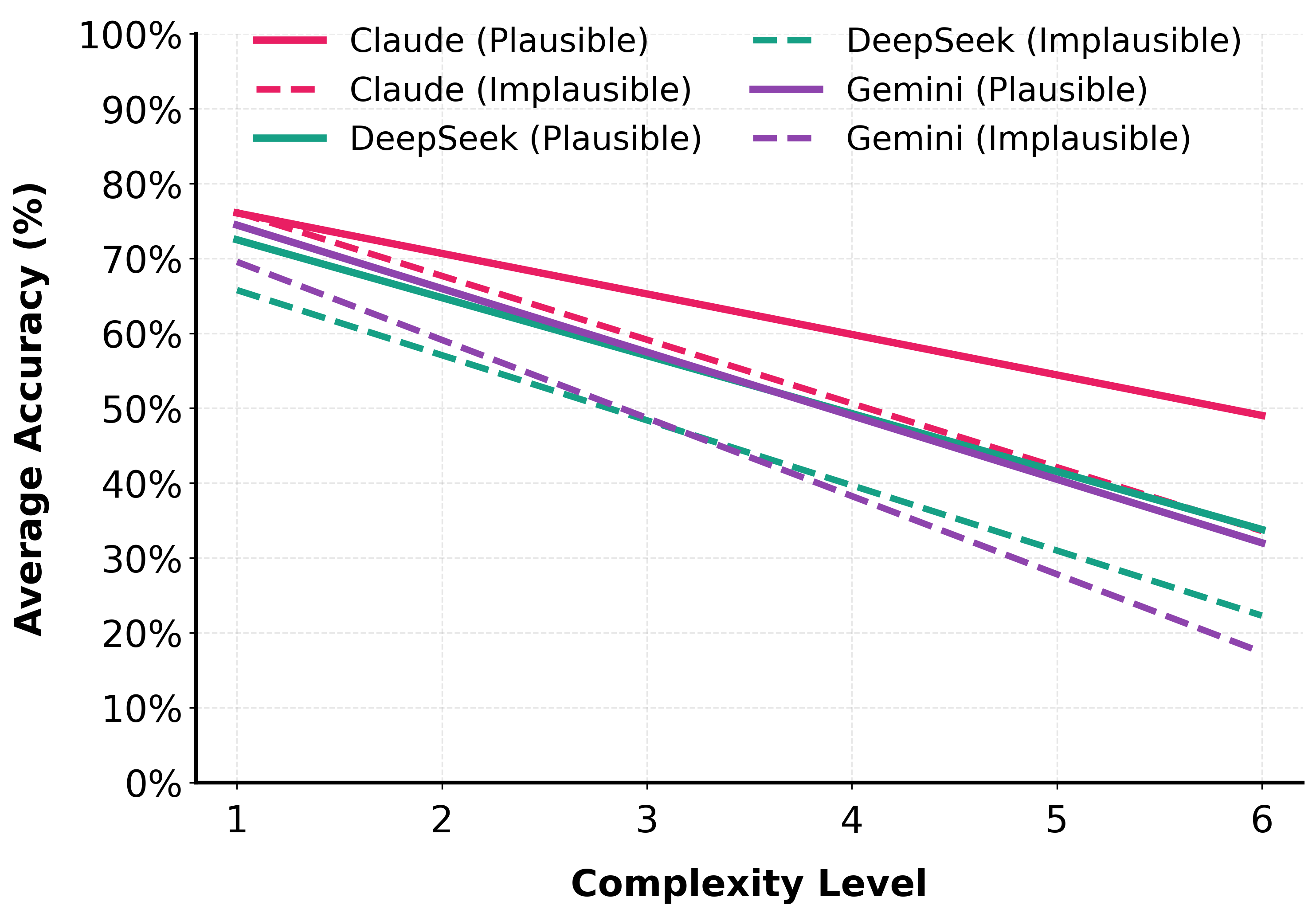}
    
    \caption{Performance degradation across complexity levels 1-6 for non-thinking models. Linear trendlines show average accuracy for plausible (solid) and implausible (dashed) sentences. All models show significant decline from level 1 to 6 ($p<0.001$).}
    \label{fig:line_graph_simple}
\end{figure}

Figure \ref{fig:line_graph_simple} shows that all non-thinking models exhibit linear performance decline as center-embedding depth increases. Accuracy drops from $\sim$80\% at complexity level 1 to 24-52\% at level 6 ($p < 0.001$), with consistent patterns across both plausible and implausible sentences.

This degradation occurs across three distinct model families (Gemini, DeepSeek, Claude), demonstrating that center-embedded sentences reliably measure syntactic processing limits. While Claude maintains the highest absolute performance, all models show similar linear decline rather than sudden collapse, indicating progressive loss of syntactic tracking ability as structural demands increase.
%(see Appendix Figure \ref{app:violen_plot_non_think} for complete performance distributions).

\subsection{Models increasingly rely on semantic associations as complexity increases}
\label{result-2}

\begin{figure}[t]
    \centering
    \includegraphics[width=0.91\columnwidth]{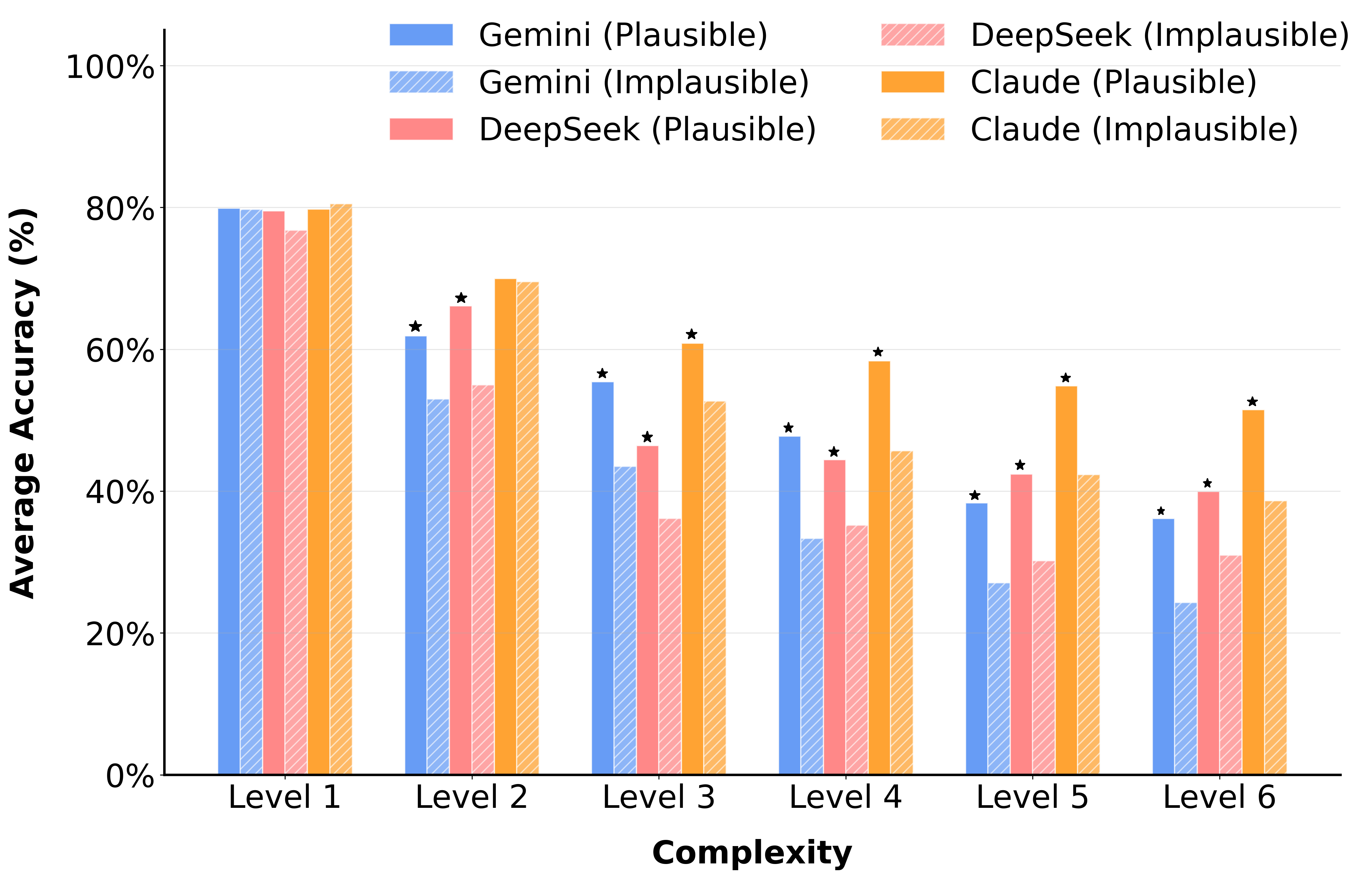}

    \caption{Average accuracy by complexity level for plausible vs.\ implausible sentences (non-thinking models). Asterisks indicate statistically significant differences between subsets at that complexity level ($p<0.0083$). The widening gap shows models increasingly rely on semantic associations rather than structural analysis.}
    \label{fig:assoc_vs_non_bars_non_think}
\end{figure}

Figure \ref{fig:assoc_vs_non_bars_non_think} shows model accuracy on plausible versus implausible sentences at each complexity level. At levels 1-2, models perform similarly on both plausible and implausible sentences. Starting at level 3, performance diverges: models score higher on plausible sentences, and this gap grows with each level, reaching over 9 percentage points by level 6.

All models show this pattern, but to different degrees. Across all complexity levels and question types, Claude's median performance gap between plausible and implausible sentences is 26.8 percentage points—this represents the median of 36 individual gaps (6 complexity levels × 6 question types), not just the overall averages shown in Figure \ref{fig:assoc_vs_non_bars_non_think}. DeepSeek shows the smallest gap at 14.6 percentage points, with Gemini at 22.6 (see Appendix Figure \ref{app:violen_plot_non_think}). %These subset differences are statistically significant from level 3 onward (p<0.0083, Bonferroni corrected).

This pattern reveals when models abandon structural analysis for semantic associations. At low complexity, they can track noun-verb relationships regardless of meaning. As complexity increases, they increasingly rely on semantic plausibility. The performance gap between subsets directly quantifies this reliance: identical syntactic structures should yield identical performance if models truly analyze structure over exploiting semantic patterns.

%Notably, Claude shows the largest median performance gap (26.8 percentage points), suggesting either greater sensitivity to semantic violations or more sophisticated use of semantic information, while DeepSeek's smaller gap (14.6 percentage points) may indicate more robust structural parsing capabilities or less reliance on semantic patterns.

\subsection{Semantic reliance varies by reasoning task}
\label{result-3}

\begin{figure}[t]
    \centering
    \includegraphics[width=0.99\columnwidth]{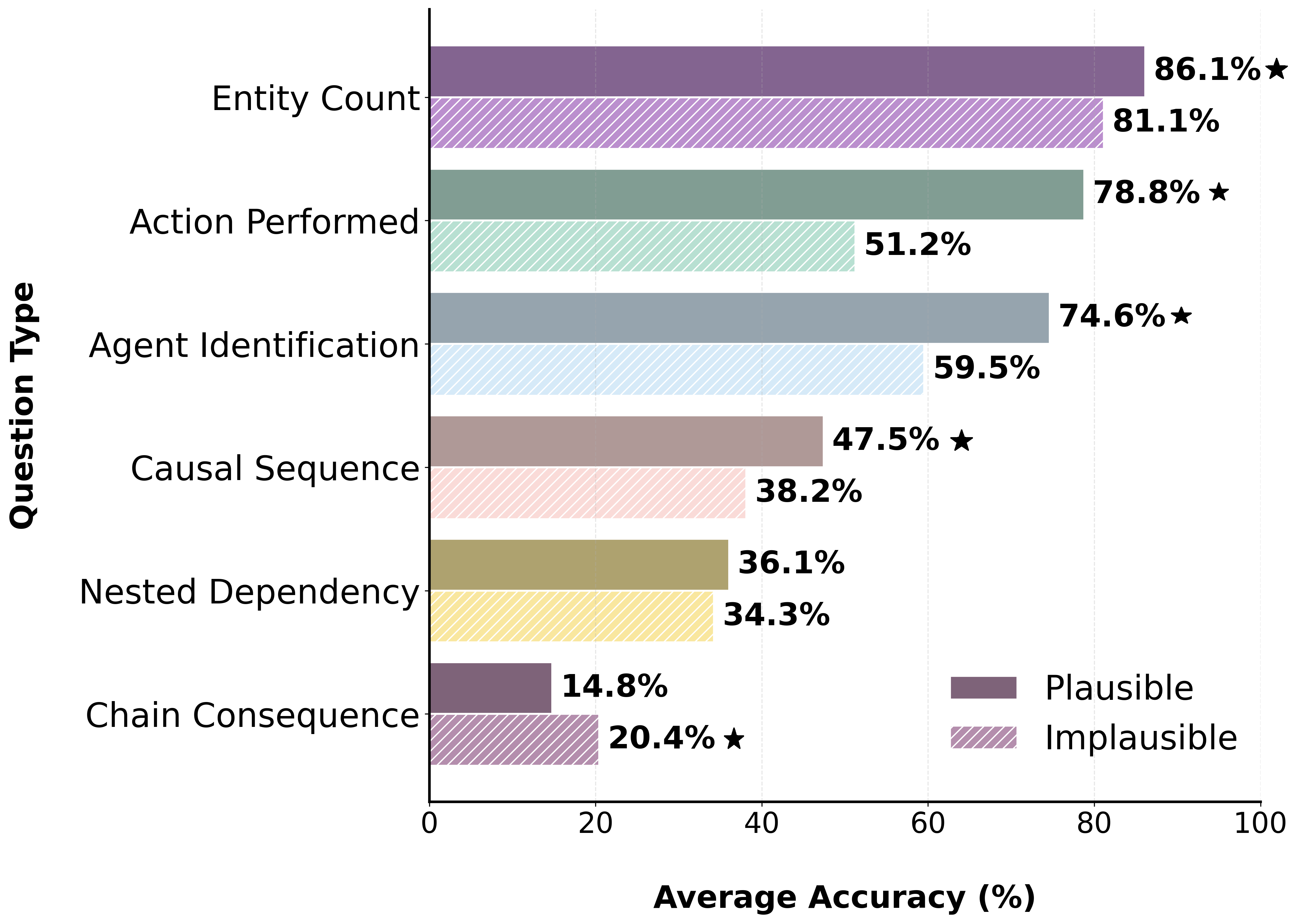}
    \caption{Average accuracy by question type for plausible vs.\ implausible sentences (non-thinking models). Asterisks mark significant differences ($p<0.0083$).}
    \label{fig:question-type-accuracy}
\end{figure}

Figure \ref{fig:question-type-accuracy} reveals a clear performance hierarchy across question types, with semantic plausibility affecting each differently. Entity counting is easiest and shows minimal plausibility effects (5.0 point gap), followed by basic comprehension tasks where plausibility provides substantial advantages: action performed (``What did X do?'': 27.6 point gap, 78.8\% vs.\ 51.2\%) and agent identification (``Who did Y to X?'': 15.1 points, 74.6\% vs.\ 59.5\%). Nested dependency questions (``What did the entity that was Y'd do?'') show almost no plausibility effect (1.8 points), while complex reasoning tasks prove hardest overall.

Notably, chain consequence questions (``What is the consequence of X's involvement?'') reverse this pattern, with implausible sentences outperforming plausible ones (20.4\% vs.\ 14.8\%. This pattern holds across all levels (Appendix Figure \ref{app:question-type-accuracy-chain-radar}). Here, semantic familiarity actively harms performance: models follow plausible associations to wrong answers rather than trace actual causal chains.

These patterns show that semantic associations aid surface comprehension but may impair complex reasoning. The questions effectively separate basic retrieval (action/agent), structural processing (entity counting, nested dependency), and multi-step reasoning (causal reasoning, chain consequence), revealing precisely when models rely on meaning versus structure.

\subsection{Reasoning models improve accuracy}

\label{result-4}
Table \ref{tab:claude_think_vs_non_comparison} demonstrates that reasoning models substantially improve performance across all complexity levels. For Claude with thinking enabled at complexity level 6, performance jumps from 18.0\% to 58.8\% (+40.8 points) on plausible sentences and from 15.6\% to 60.8\% (+45.2 points) on implausible sentences for hard questions.

For Claude, reasoning yields larger gains on implausible sentences, reducing semantic dependency: at complexity level 6, implausible sentences gain 48.9 points on easy (vs.\ +13.7 for plausible) and 45.2 points on hard (vs.\ +40.8) questions. In contrast, Gemini and DeepSeek show larger gains on plausible sentences (Appendix Tables \ref{app:gemini_think_vs_non_comparison} and \ref{app:deepseek_think_vs_non_comparison}), indicating that the effect of reasoning on plausibility is model-specific. While reasoning improves accuracy dramatically, models still achieve only $\sim$60\% on hard questions at high complexity, suggesting that explicit reasoning helps track syntactic relationships but cannot fully overcome fundamental structural processing limitations.
%Reasoning models show larger improvements on implausible sentences, reducing semantic dependency. At complexity level 6, implausible sentences gain 48.9 percentage points on easy questions (vs.\ 13.7 for plausible), 13.6 points on medium questions (vs.\ 28.4 for plausible), and 45.2 points on hard questions (vs.\ 40.8 for plausible). Similar patterns hold for Gemini and DeepSeek (Appendix Tables \ref{app:gemini_think_vs_non_comparison} and \ref{app:deepseek_think_vs_non_comparison}). 

%Furthermore, we find that CoT helps improve performance on the implausible subset. Across all question types, we observe performance gains at complexity level 6: hard questions improve from 15.6\% to 60.8\%, medium questions from 57.4\% to 71.0\%, and easy questions from 43.0\% to 91.9\%. These improvements demonstrate that systematic reasoning enables models to better parse structural relationships independent of semantic associations.

 %The consistency of this increased performance demonstrates that iterative reasoning opportunities enhance model performance on tasks with increasing linguistic complexity while simultaneously reducing dependency on semantic plausibility.

\begin{table}[!t]
\centering
\LARGE
\resizebox{1.0\columnwidth}{!}{%
\begin{tabular}{c|l|cc|cc|cc}
\toprule
\textbf{Complexity} & \textbf{Model} & \multicolumn{2}{c|}{\textbf{Easy}} & \multicolumn{2}{c|}{\textbf{Medium}} & \multicolumn{2}{c}{\textbf{Hard}} \\
\cmidrule(lr){3-4} \cmidrule(lr){5-6} \cmidrule(lr){7-8}
\textbf{Level} & \textbf{Type} & \textbf{Plaus} & \textbf{Implaus} & \textbf{Plaus} & \textbf{Implaus} & \textbf{Plaus} & \textbf{Implaus} \\
\midrule
\multirow{2}{*}{\textbf{1}} 
& Non-Think & 99.2 & 95.4 & 95.7 & 95.0 & 44.5 & 51.2 \\
& Thinking & \cellcolor{highlight}\textbf{100.0} & \cellcolor{highlight}\textbf{99.2} & \cellcolor{highlight}\textbf{94.2} & \cellcolor{highlight}\textbf{82.5} & \cellcolor{highlight}\textbf{73.3} & \cellcolor{highlight}\textbf{58.3} \\
\midrule
\multirow{2}{*}{\textbf{2}} 
& Non-Think & 88.4 & 81.8 & 74.4 & 76.1 & 47.3 & 50.7 \\
& Thinking & \cellcolor{highlight}\textbf{99.5} & \cellcolor{highlight}\textbf{93.4} & \cellcolor{highlight}\textbf{75.0} & \cellcolor{highlight}\textbf{66.1} & \cellcolor{highlight}\textbf{78.9} & \cellcolor{highlight}\textbf{71.7} \\
\midrule
\multirow{2}{*}{\textbf{3}} 
& Non-Think & 84.1 & 66.9 & 60.3 & 61.8 & 38.3 & 29.4 \\
& Thinking & \cellcolor{highlight}\textbf{97.9} & \cellcolor{highlight}\textbf{93.4} & \cellcolor{highlight}\textbf{84.2} & \cellcolor{highlight}\textbf{73.4} & \cellcolor{highlight}\textbf{74.6} & \cellcolor{highlight}\textbf{79.2} \\
\midrule
\multirow{2}{*}{\textbf{4}} 
& Non-Think & 87.0 & 53.0 & 64.4 & 59.9 & 23.8 & 24.2 \\
& Thinking & \cellcolor{highlight}\textbf{97.4} & \cellcolor{highlight}\textbf{93.4} & \cellcolor{highlight}\textbf{93.0} & \cellcolor{highlight}\textbf{75.0} & \cellcolor{highlight}\textbf{67.7} & \cellcolor{highlight}\textbf{72.4} \\
\midrule
\multirow{2}{*}{\textbf{5}} 
& Non-Think & 83.9 & 51.0 & 57.7 & 55.6 & 23.0 & 20.5 \\
& Thinking & \cellcolor{highlight}\textbf{99.5} & \cellcolor{highlight}\textbf{93.1} & \cellcolor{highlight}\textbf{92.2} & \cellcolor{highlight}\textbf{74.5} & \cellcolor{highlight}\textbf{62.8} & \cellcolor{highlight}\textbf{65.6} \\
\midrule
\multirow{2}{*}{\textbf{6}} 
& Non-Think & 80.9 & 43.0 & 55.7 & 57.4 & 18.0 & 15.6 \\
& Thinking & \cellcolor{highlight}\textbf{94.6} & \cellcolor{highlight}\textbf{91.9} & \cellcolor{highlight}\textbf{84.1} & \cellcolor{highlight}\textbf{71.0} & \cellcolor{highlight}\textbf{58.8} & \cellcolor{highlight}\textbf{60.8} \\
\midrule
\bottomrule
\end{tabular}%
}

\caption{Claude: Thinking vs Non-Thinking performance across question difficulty levels on plausible (Plaus) and implausible (Implaus) subsets}
\label{tab:claude_think_vs_non_comparison}
\end{table}

\subsection{Reasoning traces provide evidence of systematic processing failures}
\label{result-5}

Analysis of reasoning traces exposes how thinking models fail when processing complex structures. The traces suggest several distinct failure patterns (see Appendix Tables \ref{tab:qualitative_analysis_tab}-\ref{tab:qualitative_analysis_appendix_tab_1}):

\paragraph{Semantic interference in parsing:} Models' reasoning shows they actively seek semantic coherence over syntactic accuracy. When Gemini processes ``the bicycle... orbited cycled around,'' its trace shows it selecting ``cycled around'' because ``that the bicycle... cycled around'' makes semantic sense, completely missing that ``orbited'' is the syntactically correct verb (Appendix Table \ref{tab:qualitative_analysis_tab}). This indicates that models don't just happen to fail on implausible sentences; their reasoning explicitly prioritizes meaning over structure.

\paragraph{Refusal when structure conflicts with world knowledge:} Models handle implausible results in remarkably different ways. Gemini refuses to answer when parsing yields semantically odd relationships, stating ``no one in this sentence directly reads rights to the mailman'' even though the syntactic chain clearly indicates the teacher. In contrast, DeepSeek produces the correct answer but only after 3,498 tokens of circular reasoning, while Claude methodically parses the structure and answers correctly in 601 tokens (Appendix Table \ref{tab:qualitative_analysis_appendix_tab_1}).

\paragraph{Self-induced errors through overthinking:} Perhaps most revealing, models create errors in simple tasks through excessive reasoning. On basic entity counting, both DeepSeek and Claude initially identify the correct answer (4 entities) but then spiral into doubt: ``I think I'm overcomplicating... Perhaps the answer is 3, excluding the case. I'm not sure.'' Meanwhile, Gemini shows overconfidence, declaring the task ``Elementary'' while still arriving at the wrong answer. This pattern shows that explicit reasoning can actively harm performance on straightforward tasks (Appendix Table \ref{tab:qualitative_analysis_appendix_tab_2}).

\subsection{Human performance varies unpredictably with complexity}
\label{result-6}

\begin{figure}[!t]
    \centering
    \includegraphics[width=0.99\linewidth]{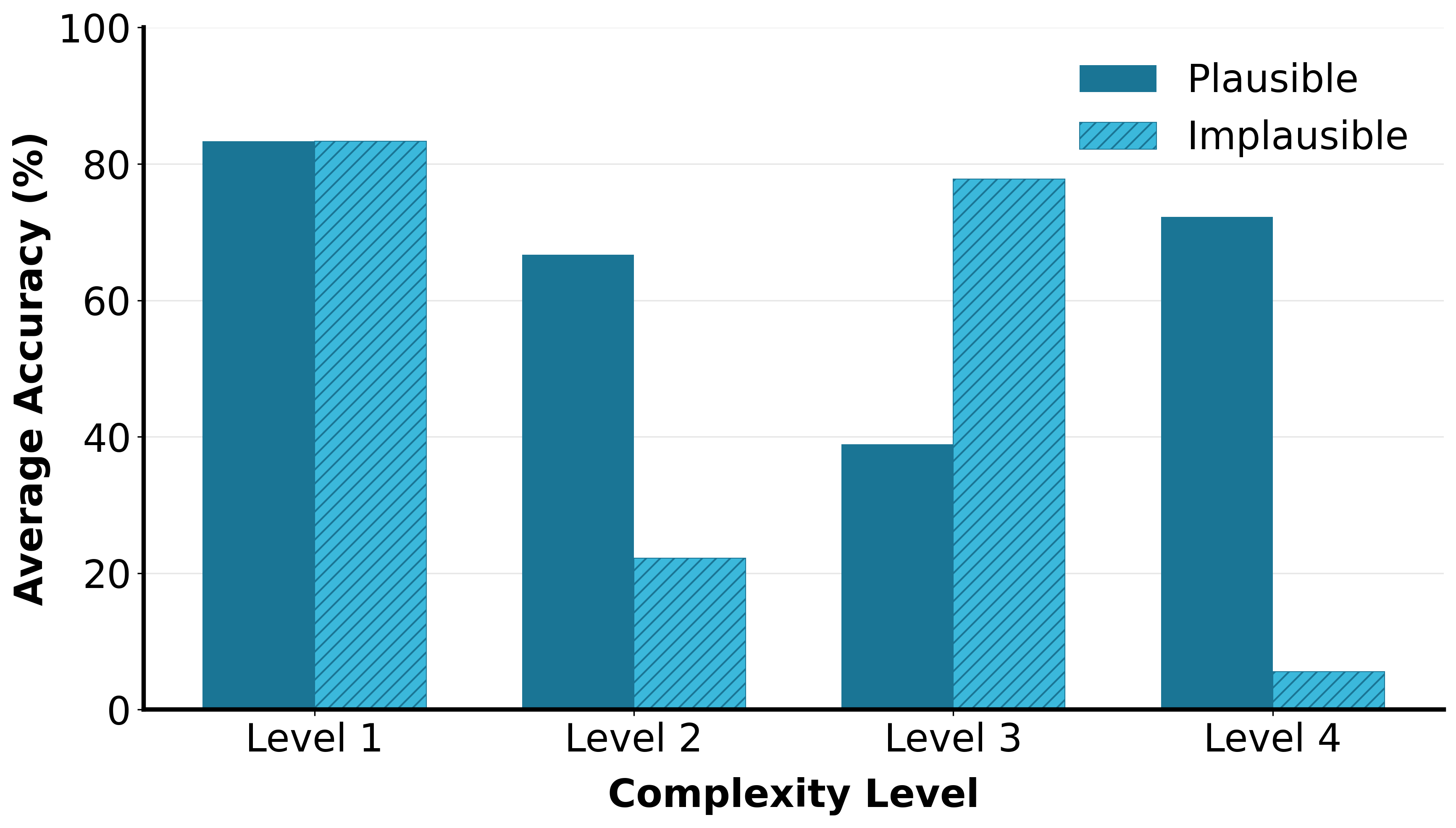}
    
    \caption{Human accuracy on plausible and implausible sentences across complexity levels 1-4.}
    
    \label{fig:human_performance}
\end{figure}

Figure \ref{fig:human_performance}  reveals marked differences between human and model processing patterns. Unlike models, which show consistent semantic advantages that grow with complexity, humans exhibit variable effects: equal performance on both subsets at level 1, plausible advantage at levels 2 and 4, but implausible advantage at level 3. 

Human accuracy also differs from model patterns in absolute performance. While models show steady degradation with complexity, human performance fluctuates: starting at 83\% (level 1), dropping to 67\% (level 2) and 39\% (level 3), then rising to 72\% (level 4) on plausible sentences. This non-monotonic pattern contrasts sharply with models' linear decline. This suggests humans may engage different processing strategies when semantic cues conflict with structural demands, potentially paying closer attention to syntax when meaning provides no shortcuts.

\section{Related Works}

\paragraph{Linguistic Complexity and Performance Degradation}
While recent work shows language models struggle with increasing task complexity \cite{qi2024quantifyinggeneralizationcomplexitylarge, kang-etal-2024-empirical, hua-etal-2025-inductionbench}, these studies focus on symbolic reasoning, mathematical operations \cite{dziri2023faithfatelimitstransformers}, or puzzle-solving \cite{estermann2024puzzlesbenchmarkneuralalgorithmic}. Such tasks conflate domain-specific knowledge with structural processing, obscuring whether failures are linguistic or task-specific. Center-embedded sentences offer a pure test of recursive syntactic processing established through decades of psycholinguistic research \cite{Chomsky1963-CHOITT-2, Gibson1998-xu}, scaling complexity through clause nesting without requiring external knowledge.

\paragraph{Center-Embedding in Computational Linguistics} 
Center-embedded sentences have long served as a benchmark for syntactic processing, yet existing computational resources remain limited. Early work by \citet{wilcox-etal-2019-hierarchical} tested hierarchical representations with 28 plausible/implausible sentence pairs, but only at single embedding depth. SyntaxGym \cite{gauthier-etal-2020-syntaxgym} and \citet{hu-etal-2020-systematic} reuse these single-level items. GECS \cite{carslaw2025automaticextractionclausalembedding} extracts naturally-occurring embeddings from web text but lacks systematic complexity scaling or plausibility control. While \citet{hardt2025sparks} tested models on one to four embedding levels with noun-verb action questions, their sentences ignored plausibility. Studies comparing human and model processing \cite{wilcox-etal-2021-targeted, Lakretz_2021, lakretz-etal-2022-transformers} find both struggle with deeper embeddings but differently: humans maintain above-chance performance while LSTM accuracy collapses, and even strong Transformers fail at long-range dependencies. These limitations motivate our systematic manipulation of both complexity and plausibility with diverse question types.

\paragraph{Semantic Interference in Language Processing}
Extensive evidence shows both humans and models rely on semantic plausibility, though differently. Humans employ ``good enough'' processing strategies \cite{ferreira2002good, frances2024good}, while models show systematic biases toward training knowledge. \citet{wilcox-etal-2019-hierarchical} found models assign higher surprisal to implausible pairings, correctly tracking dependencies but revealing semantic expectations. Studies using counterfactual contexts expose stronger biases: Fakepedia shows GPT-4 ignores false passages to give memorized answers 99\% of the time \cite{monea-etal-2024-glitch}, while entity substitution frameworks show models prefer parametric over contextual knowledge \cite{longpre-etal-2021-entity, neeman2022disentqadisentanglingparametriccontextual}. Additionally, recent work demonstrates plausibility's diagnostic power beyond syntax:  \citet{palta-etal-2024-plausibly} use plausibility ratings to identify quality issues in existing commonsense benchmarks where semantic expectations conflict with gold labels. 
However, these test binary memory-context conflicts. Our work identifies the \textit{continuous} transition: by scaling syntactic complexity while controlling plausibility, we pinpoint when models shift from structural analysis to semantic shortcuts.

\section{Conclusion}

We introduced \corpusname{}, a dataset of 9,720 test instances varying complexity, reasoning demands, and semantic plausibility. Models show consistent performance degradation as complexity increases, while also increasingly relying on meaning over structure, with performance gaps reaching 25+ percentage points. Interestingly, semantic plausibility backfires on complex reasoning, where familiar patterns mislead models when structural analysis is required. Reasoning models improve accuracy but their traces provide evidence of systematic failures including prioritizing semantic coherence, refusing implausible answers, and overthinking simple tasks. These patterns contrast sharply with humans' inconsistent semantic effects. By revealing precisely when and how models abandon structural analysis for semantic shortcuts, \corpusname{} enables informed decisions about model deployment in domains requiring genuine syntactic understanding.

\section*{Limitations}

We structure our limitations section as arguments and counterarguments, inspired by \citet{balepur-etal-2025-best}:

\paragraph{The semantic manipulation targets more than plausibility:} 
Our implausible sentences simultaneously manipulate collocational strength and verb-noun associations, this is intentional, not confounding. Semantic plausibility cannot be isolated from these factors because they collectively constitute what makes language plausible. As demonstrated by \citet{lapata-etal-1999-determinants}, co-occurrence frequency (which captures collocational relationships) is the strongest predictor of plausibility judgments. By breaking familiar associations (doctors prescribing → doctors delivering mail), we create the semantic disruption needed to test when models abandon syntax for semantics. We maintain domain consistency within sentences (all vehicles, all professions) and keep syntactic structure identical across matched pairs, ensuring that performance differences reflect reliance on semantic cues.

\paragraph{The semantic manipulation is too artificial:} Creating sentences where ``doctors deliver mail'' and ``mailmen prescribe medicine'' might seem contrived, but these actions remain perfectly possible—they simply violate our world knowledge and expectations \cite{wu-etal-2024-reasoning}. By preserving grammatical correctness while disrupting typical semantic associations, we force models to choose between structural and semantic processing. The fact that this manipulation consistently affects performance reveals how deeply models depend on semantic plausibility, a finding with direct implications for domains where semantic cues may be unreliable, such as processing figurative language, analyzing texts from different time periods, or handling adversarial inputs \cite{zhang-etal-2024-gome,madhusudan-etal-2025-fine,wallace-etal-2019-trick}.

\paragraph{This is just another syntax benchmark:} While center-embedded sentences might seem like an artificial linguistic construction, they represent a fundamental test of recursive structural processing that appears across natural language. Unlike toy puzzles designed solely for benchmarking, these structures have been studied for decades in psycholinguistics precisely because they reveal how humans and systems handle hierarchical dependencies \cite{Miller1963-MILFMO, hudson1996difficulty}. Our systematic manipulation of semantic plausibility transforms this classical paradigm into a diagnostic that can identify when any language system shifts from structural analysis to pattern matching.

\paragraph{You can't definitively prove models are or aren't ``parsing'':} We acknowledge that our comprehension questions don't directly test syntactic parsing in the formal sense. However, this is precisely why we designed matched sentence pairs: if models were truly building syntactic representations, they should perform identically on structures that differ only in semantic plausibility \cite{mueller-etal-2024-context}. The systematic performance gaps we observe provide strong evidence that models rely on semantic associations rather than structural analysis, regardless of their internal representations.

\paragraph{Real language use doesn't involve such complex embeddings:} Corpus studies confirm that deeply nested structures rarely appear in natural text \cite{Karlsson2007}. But this is exactly why they're valuable for evaluation: they test whether models have learned generalizable principles of syntactic structure or merely memorized common patterns. Just as stress tests in engineering reveal failure modes through extreme conditions, our complexity scaling identifies the precise point where models abandon whatever structural processing they might possess for semantic shortcuts.

\paragraph{Why not use the same entities doing implausible things?} One might wonder why we don't simply keep the same entities from plausible sentences and have them perform implausible actions (\textit{cats barking, dogs meowing}). While this alternative seems straightforward, it's surprisingly difficult to execute. Consider the constraints: we need actions that \textit{clearly} violate semantic expectations, maintaining grammatical requirements (correct transitive/intransitive patterns), avoiding temporal violations, and working across 2-7 entity chains. Finding sufficient verbs meeting these criteria for each entity becomes intractable. Our approach—circular verb swapping across animals, occupations, and vehicles—provides clear plausibility violations (skunks roaring, mailmen prescribing medicine, bicycles orbiting) while maintaining all syntactic constraints. These actions remain \underline{physically possible} but violate our world knowledge and expectations, creating the semantic disruption we need. The entities are merely vehicles for testing whether models maintain syntactic parsing when semantic cues are removed. What matters is the within-model performance gap on identical syntactic structures, which directly quantifies reliance on semantic shortcuts over structural analysis.

\paragraph{Humans only went to complexity level 4:} Indeed, human participants showed increasing difficulty with our sentences, and we limited testing to level 4 for practical reasons. However, this actually strengthens our findings: even at levels where humans can still process these structures (albeit with effort), they show qualitatively different patterns from models. The inconsistent semantic effects in humans align with research showing that human sentence processing often relies on ``good enough'' representations rather than complete structural analysis \cite{ferreira2002good, frances2024good}, contrasting with models' systematic biases that consistently favor semantic plausibility.

\bibliography{custom.bib}
\clearpage
\appendix
\onecolumn
\section{Appendix}
\subsection{Dataset}
\subsubsection{Nouns and Verbs in \corpusname{}}
\label{sec: nouns_and_verbs}
\begin{table*}[h!]
\centering
\renewcommand{\arraystretch}{1.3}
\resizebox{\textwidth}{!}{
\begin{tabular}{@{}p{2.5cm}p{3.5cm}p{6.5cm}p{6.5cm}@{}}
\toprule
\textbf{Category} & \textbf{Subcategory} & \textbf{Included Nouns} & \textbf{Excluded Nouns} \\
\midrule

\multirow{5}{*}{Animals} & Birds & 
Eagle, Sparrow, Crow, Pigeon, Parrot, Chicken, Duck, Owl, Goose, Vulture & 
Blue Jay, Robin, Cardinal, Hawk, Hummingbird, Dove, Finch, Raven, Woodpecker, Wren, Ostrich, Parakeet, Seagull, Flamingo, Mockingbird, Penguin, Falcon, Black Bird, Oriole, Swan, Canary, Turkey, Swallow, Chickadee, Crane, Emu, Grackle, Heron, Jaybird, Peacock, Pelican, Red Bird, Starling, Stork \\

& Fish & 
Salmon, Tuna, Goldfish, Trout, Shark, Whale, Piranha, Starfish & 
Catfish, Bass, Cod, Carp, Perch, Tilapia, Flounder, Sword Fish, Clown Fish, Blue Gill, Pike, Crappie, Guppy, Minnow, Halibut, Grouper, Snapper, Sun Fish, Koi, Puffer Fish, Angel Fish, Barracuda, Blowfish, Blue, Dolphin, Mackerel, Marlin, Pollock, Sturgeon, Walleye \\

& Four-Footed Animals & 
Dog, Cat, Horse, Lion, Cow, Tiger, Bear, Elephant, Deer, Pig, Giraffe, Mouse, Sheep, Goat, Rat, Wolf, Zebra, Donkey, Rabbit, Raccoon, Squirrel, Coyote, Cougar, Moose, Cheetah, Rhinoceros, Fox, Mule, Hippopotamus, Beaver, Camel, Ferret, Frog, Jaguar, Lamb, Leopard, Lizard, Llama, Skunk & 
Opossum, Hamster, Elk \\

& Snakes & 
Rattlesnake, Cobra, Python, Anaconda, Viper, Black Mamba & 
Garden, Boa Constrictor, Copperhead, Water Moccasin, Black, Asp, Cottonmouth, Coral, King, Rat, Diamondback, Green, Corn Snake, Grass, Mamba, Racer \\

& Insects & 
Ant, Fly, Bee, Spider, Beetle, Mosquito, Cockroach, Wasp, Lady Bug, Butterfly, Grasshopper, Moth, Gnat, Cricket, Caterpillar, Worm, Centipede, Termite, Praying Mantis, Bug, Dragonfly & 
Flea, Stink Bug, Hornet, Millipede, Rolly Polly, Tick, Yellow Jacket \\

\midrule

Occupations &  & 
Doctor, Lawyer, Teacher, Nurse, Police Officer, Firefighter, Accountant, Dentist, Engineer, Plumber, Carpenter, Salesperson, Secretary, Cashier, Mechanic, Professor, Electrician, Chef, Scientist, Clerk, Banker, Actor, Truck Driver, Mailman, Artist, Athlete, Attorney, Bus Driver, CEO, Construction Worker, Garbage Man, Janitor, Judge, Manager, Musician, Pilot, Politician, Programmer, Surgeon, Veterinarian, Waiter, Writer & 
None \\

Vehicles &  & 
Car, Truck, Motorcycle, Airplane, Bicycle, Bus, SUV, Boat, Train, Van, Jeep, Tank, Sedan, Tractor, Scooter, RV, Taxi, ATV, Ambulance, Convertible, Golf Cart, Helicopter, Pickup Truck, Ship, Sled, Snowmobile, Sports Car & 
Semi, Minivan, Station Wagon, Wagon \\

\bottomrule
\end{tabular}
}
\caption{Included and excluded nouns from \citet{VANOVERSCHELDE2004289} across categories}
\label{appendix:noun-inclusion}
\end{table*}

\begin{table*}[h!]
\centering
\renewcommand{\arraystretch}{1.3}
\resizebox{0.8\textwidth}{!}{
\begin{tabular}{@{}p{3cm}p{14cm}@{}}
\toprule
\textbf{Category} & \textbf{Nouns} \\
\midrule

Animals & 
Eagle, Sparrow, Crow, Pigeon, Parrot, Chicken, Duck, Owl, Goose, Vulture, Salmon, Tuna, Goldfish, Trout, Shark, Whale, Piranha, Starfish, Dog, Cat, Horse, Lion, Cow, Tiger, Bear, Elephant, Deer, Pig, Giraffe, Mouse, Sheep, Goat, Rat, Wolf, Zebra, Donkey, Rabbit, Raccoon, Squirrel, Coyote, Cougar, Moose, Cheetah, Rhinoceros, Fox, Mule, Hippopotamus, Beaver, Camel, Ferret, Frog, Jaguar, Lamb, Leopard, Lizard, Llama, Skunk, Rattlesnake, Cobra, Python, Anaconda, Viper, Black Mamba, Ant, Fly, Bee, Spider, Beetle, Mosquito, Cockroach, Wasp, Ladybug, Butterfly, Grasshopper, Moth, Gnat, Cricket, Caterpillar, Worm, Centipede, Termite, Praying Mantis, Bug, Dragonfly, Pelican \\

\midrule

Occupations & 
Doctor, Lawyer, Teacher, Nurse, Police Officer, Firefighter, Accountant, Dentist, Engineer, Plumber, Carpenter, Salesperson, Secretary, Cashier, Mechanic, Professor, Electrician, Chef, Scientist, Clerk, Banker, Actor, Truck Driver, Mailman, Artist, Athlete, Attorney, Bus Driver, CEO, Construction Worker, Garbage Man, Janitor, Judge, Manager, Musician, Pilot, Politician, Programmer, Surgeon, Veterinarian, Waiter, Writer \\

\midrule

Vehicles & 
Car, Truck, Motorcycle, Airplane, Bicycle, Bus, SUV, Boat, Train, Van, Jeep, Tank, Sedan, Tractor, Scooter, RV, Taxi, ATV, Ambulance, Convertible, Golf Cart, Helicopter, Pickup Truck, Ship, Sled, Snowmobile, Sports Car \\

\midrule

GPT-4 Extensions & 
Lioness, Gazelle, Calf, Cub, Duckling, Kitten, Puppy, Bird, Snake, Mongoose, Child, Baby, Infant, Parent, Toddler, Boy, Girl, Man, Woman, Teenager, Kid, Adult, Youth, Journalist, Reporter, Director, Editor, Photographer, Detective, Investigator, Researcher, Student, Supervisor, Inspector, Principal, Hunter, Gatherer, Villager, Resident, Tourist, Passenger, Driver, Maid, Host, Guest, Predator, Prey, Scavenger, Beast, Creature, Animal, Minivan, Submarine, Spaceship, Drone, Hound, Dove, Flagged, Architect, Mayor, Dean, Cyclist, Crab, Seagull, Paper, Publisher, Fish, Letter, Thief, Chief, Hawk, Zookeeper, Patient, Suitcase, Porter, Park Ranger, Bat, Ranger, Neighbor, Heron, Hiker, Guide, Owner, Postman, Policeman, Homeowner, Vet, Octopus, Diver, Captain, Farmer, Sheriff, Nanny, Landlord, Forester, Jogger, Coach, Wife, Mother, Security Guard, Grandfather, Pet Owner, Tenant, Trapper, Conservationist, Gardener, Cheese, Shopkeeper, Superintendent, Animal Rights Activist, Falconer, Babysitter, Grandmother, Game Warden, Zoologist, Biologist, Father, Pest Controller, Flower, Nut, Naturalist, Intern, Skateboard, Seal, Dolphin, Fisherman, Coast Guard, Note, Investor, President, Pearl, Oyster, Chip, Message, Chancellor, Leaf, Chameleon, Apple, Storm, Citizen, Baby Bird \\

\bottomrule
\end{tabular}
}
\caption{Complete list of nouns in the plausible dataset by category}
\label{appendix: all-noun-categories}
\end{table*}

\begin{table}[htbp]
\centering
\small
\resizebox{0.8\textwidth}{!}{%
\begin{tabular}{lll}
\toprule
\textbf{Category} & \textbf{Noun} & \textbf{Verbs (Transitive / Intransitive)} \\
\midrule
\multirow{6}{*}{Animals} 
& parrot & mimicked / chattered \\
& horse & neighed at, galloped toward / neighed, galloped \\
& elephant & trumpeted at, charged at / trumpeted, stomped \\
& bee & buzzed at, stung / buzzed, swarmed \\
& cobra & hissed at, reared at / hissed, reared, coiled \\
\midrule
\multirow{6}{*}{Professions}
& doctor & prescribed medicine to, examined / prescribed medicines, diagnosed \\
& lawyer & cross-examined, defended in court / argued, objected \\
& police officer & arrested, read rights to / patrolled, responded \\
& firefighter & rescued, carried out / extinguished fire, rescued \\
& judge & sentenced, ruled against / presided, gaveled \\
\midrule
\multirow{6}{*}{Vehicles}
& truck & jackknifed into, towed / jackknifed \\
& motorcycle & lane-split past, wheelied at / wheelied \\
& airplane & air-dropped on, strafed / taxied \\
& bicycle & pedaled past, cycled around / pedaled \\
& ambulance & blared the siren at / blared the siren \\
\bottomrule
\end{tabular}%
}
\caption{Sample nouns and associated verbs by category generated by \textit{claude-sonnet-4-20250514} for implausible sentence generation}
\label{app:entity_verb_samples}
\end{table}

\clearpage

\subsubsection{Sentence Generation - Plausible}
\label{sec: sentence_generation}

\paragraph{Complexity Level Specifications}
The system dynamically generated prompts based on complexity level, where each level follows a systematic pattern of increasing structural complexity. Each complexity level is defined by the number of entities, verbs, and embedded ``that'' clauses:

\begin{description}[leftmargin=0pt]
\item[\textbf{Complexity Level 1}] (2 \textit{entities}, 2 \textit{verbs}, 1 ``that'' clause)
\begin{itemize}[nosep]
    \item \textit{structure:} \texttt{[A] that [B] [verb1] [verb2]}
    \item \textit{semantic\_rule:} B performs verb1 TO A, then A performs verb2
    \item \textit{example:} ``The mouse that the cat chased escaped."
    \item \textit{flow:} CAT chased MOUSE $\rightarrow$ MOUSE escaped
\end{itemize}

\item[\textbf{Complexity Level 2}] (3 \textit{entities}, 3 \textit{verbs}, 2 ``that'' clauses)
\begin{itemize}[nosep]
    \item \textit{structure:} \texttt{[A] that [B] that [C] [verb1] [verb2] [verb3]}
    \item \textit{semantic\_rule:} C performs verb1 TO B, B performs verb2 TO A, A performs verb3
    \item \textit{example:} ``The fly that the spider that the bird saw stalked buzzed."
    \item \textit{flow:} BIRD saw SPIDER $\rightarrow$ SPIDER stalked FLY $\rightarrow$ FLY buzzed
\end{itemize}

\item[\textbf{Complexity Level 3}] (4 \textit{entities}, 4 \textit{verbs}, 3 ``that'' clauses)
\begin{itemize}[nosep]
    \item \textit{structure:} \texttt{[A] that [B] that [C] that [D] [verb1] [verb2] [verb3] [verb4]}
    \item \textit{semantic\_rule:} D performs verb1 TO C, C performs verb2 TO B, B performs verb3 TO A, A performs verb4
    \item \textit{example:} ``The worm that the bird that the cat that the dog chased saw ate died."
    \item \textit{flow:} DOG chased CAT $\rightarrow$ CAT saw BIRD $\rightarrow$ BIRD ate WORM $\rightarrow$ WORM died
\end{itemize}

\item[\textbf{Complexity Level 4}] (5 \textit{entities}, 5 \textit{verbs}, 4 ``that'' clauses)
\begin{itemize}[nosep]
    \item \textit{structure:} \texttt{[A] that [B] that [C] that [D] that [E] [verb1] [verb2] [verb3] [verb4] [verb5]}
    \item \textit{semantic\_rule:} E performs verb1 TO D, D performs verb2 TO C, C performs verb3 TO B, B performs verb4 TO A, A performs verb5
    \item \textit{example:} ``The mouse that the cat that the dog that the owner that the neighbor called trained chased caught squeaked."
    \item \textit{flow:} NEIGHBOR called OWNER $\rightarrow$ OWNER trained DOG $\rightarrow$ DOG chased CAT $\rightarrow$ CAT caught MOUSE $\rightarrow$ MOUSE squeaked
\end{itemize}

\item[\textbf{Complexity Level 5}] (6 \textit{entities}, 6 \textit{verbs}, 5 ``that'' clauses)
\begin{itemize}[nosep]
    \item \textit{structure:} \texttt{[A] that [B] that [C] that [D] that [E] that [F] [verb1] [verb2] [verb3] [verb4] [verb5] [verb6]}
    \item \textit{semantic\_rule:} F performs verb1 TO E, E performs verb2 TO D, D performs verb3 TO C, C performs verb4 TO B, B performs verb5 TO A, A performs verb6
    \item \textit{example:} ``The ant that the spider that the lizard that the snake that the hawk that the hunter saw spotted followed grabbed saw crawled."
    \item \textit{flow:} HUNTER saw HAWK $\rightarrow$ HAWK spotted SNAKE $\rightarrow$ SNAKE followed LIZARD $\rightarrow$ LIZARD grabbed SPIDER $\rightarrow$ SPIDER saw ANT $\rightarrow$ ANT crawled
\end{itemize}

\item[\textbf{Complexity Level 6}] (7 \textit{entities}, 7 \textit{verbs}, 6 ``that'' clauses)
\begin{itemize}[nosep]
    \item \textit{structure:} \texttt{[A] that [B] that [C] that [D] that [E] that [F] that [G] [verb1] [verb2] [verb3] [verb4] [verb5] [verb6] [verb7]}
    \item \textit{semantic\_rule:} G performs verb1 TO F, F performs verb2 TO E, E performs verb3 TO D, D performs verb4 TO C, C performs verb5 TO B, B performs verb6 TO A, A performs verb7
    \item \textit{example:} ``The crumb that the ant that the spider that the lizard that the snake that the hawk that the eagle observed followed chased startled carried dropped rolled."
    \item \textit{flow:} EAGLE observed HAWK $\rightarrow$ HAWK followed SNAKE $\rightarrow$ SNAKE chased LIZARD $\rightarrow$ LIZARD startled SPIDER $\rightarrow$ SPIDER carried ANT $\rightarrow$ ANT dropped CRUMB $\rightarrow$ CRUMB rolled
\end{itemize}
\end{description}

\vspace{0.3cm}
\paragraph{System Prompt}
\vspace{0.3cm} We employed \textit{GPT-4-0613} with temperature 0.7. The  system prompt used was:

\begin{quote}
\small
You are a linguistics expert specializing in center-embedded sentences.

\# COMPLEXITY LEVEL \{complexity\_level\} SPECIFICATIONS
\begin{itemize}
\item Required entities: \{entities\}
\item Required verbs: \{verbs\}
\item Required ``that'' clauses: \{complexity\_level\}
\item Embedding depth: \{complexity\_level\} levels
\end{itemize}

\# GRAMMATICAL STRUCTURE\\
Pattern: \{structure\}

\# SEMANTIC FLOW RULES (CRITICAL)\\
Core Rule: \{semantic\_rule\}\\
Example: \{example\}\\
Action Flow: \{flow\}

\# TEMPORAL CONSISTENCY RULES (MANDATORY)
\begin{enumerate}
\item Actions must follow logical temporal sequence
\item Dead entities CANNOT perform subsequent actions
\item Caught/trapped/seized entities CANNOT act on other entities
\item Eaten entities CANNOT perform actions after being consumed
\item Actions must respect cause-and-effect relationships
\item No temporal paradoxes or impossibilities allowed
\end{enumerate}

\# STRUCTURAL REQUIREMENTS
\begin{enumerate}
\item Use ``that'' as relative pronoun for ALL embeddings
\item Verbs appear in REVERSE order of entity introduction
\item Last entity introduced performs FIRST action
\item First entity performs FINAL action (must be intransitive)
\item Each ``that'' clause introduces exactly ONE entity
\item Actions flow from innermost clause outward
\end{enumerate}

\# SEMANTIC PLAUSIBILITY CONSTRAINTS\\
Predator-Prey Relationships:
\begin{itemize}
\item Must reflect realistic natural hierarchies
\item Size/strength differences must be logical
\item Hunting behaviors must be species-appropriate
\end{itemize}

Professional Relationships:
\begin{itemize}
\item Authority structures must be realistic
\item Professional interactions must be plausible
\item Skills must match occupations
\end{itemize}

Physical Capabilities:
\begin{itemize}
\item Actions must match entity capabilities
\item Environmental constraints must be respected
\item Biological limitations must be observed
\end{itemize}

\# MEMORY AID: NESTED ACTION PRINCIPLE\\
Think of Russian dolls opening from inside out:
\begin{itemize}
\item Innermost doll (last entity) acts first
\item Each outer doll (entity) acts on the result
\item Outermost doll (first entity) performs final action
\item Each action must be temporally possible given previous actions
\end{itemize}

\# QUALITY REQUIREMENTS
\begin{itemize}
\item Sentences must be grammatically perfect
\item Semantic relationships must be crystal clear
\item No ambiguous temporal references
\item All actions must be logically sequenced
\item Maintain subject-verb agreement throughout
\end{itemize}
\end{quote}

\paragraph{User Prompt}
The user prompt used was:

\begin{quote}
\vspace{-2mm}
\small
Generate 30 unique center-embedded sentences at complexity level \{complexity\_level\}.

\# STRICT REQUIREMENTS
\begin{itemize}
\item Exactly \{entities\} different entities
\item Exactly \{verbs\} verbs  
\item Exactly \{complexity\_level\} ``that'' clauses
\item Perfect temporal consistency - NO temporal violations
\item Semantically plausible relationships
\item Grammatically correct structure
\end{itemize}

\# ENTITY CATEGORIES

Animals:\\
eagle, sparrow, crow, pigeon, parrot, chicken, duck, owl, goose, vulture, salmon, tuna, goldfish, trout, shark, whale, piranha, starfish, dog, cat, horse, lion, cow, tiger, bear, elephant, deer, pig, giraffe, mouse, sheep, goat, rat, wolf, zebra, donkey, rabbit, raccoon, squirrel, coyote, cougar, moose, cheetah, rhinoceros, fox, mule, hippopotamus, beaver, camel, ferret, frog, jaguar, lamb, leopard, lizard, llama, skunk, rattlesnake, cobra, python, anaconda, viper, black mamba, ant, fly, bee, spider, beetle, mosquito, cockroach, wasp, ladybug, butterfly, grasshopper, moth, gnat, cricket, caterpillar, worm, centipede, termite, praying mantis, bug, dragonfly

People (Occupations):\\
doctor, lawyer, teacher, nurse, police officer, firefighter, accountant, dentist, engineer, plumber, carpenter, salesperson, secretary, cashier, mechanic, professor, electrician, chef, scientist, clerk, banker, actor, truck driver, mailman, artist, athlete, attorney, bus driver, CEO, construction worker, garbage man, janitor, judge, manager, musician, pilot, politician, programmer, surgeon, veterinarian, waiter, writer

Vehicles:\\
car, truck, motorcycle, airplane, bicycle, bus, SUV, boat, train, van, jeep, tank, sedan, tractor, scooter, RV, taxi, ATV, ambulance, convertible, golf cart, helicopter, pickup truck, ship, sled, snowmobile, sports car

\# CRITICAL REMINDERS
\begin{itemize}
\item After an entity is caught/killed/eaten, it CANNOT perform actions
\item Predator-prey relationships must be biologically accurate  
\item Professional hierarchies must be realistic
\item Actions must follow logical temporal sequence
\item Each sentence must tell a coherent, plausible story
\end{itemize}

\# OUTPUT FORMAT\\
Number each sentence 1-30, one per line:\\
1. [sentence]\\
2. [sentence]\\
...\\
30. [sentence]

\end{quote}

\clearpage

\subsubsection{Sentence Generation - Implausible}
\label{sec:algorithm}

\textbf{Algorithm Implementation}

The following algorithm implements the circular verb swapping procedure described in Section \ref{sec: sentence-implausible}:

\begin{algorithm}[H]
\caption{Circular Verb Swapping for Implausible Sentence Generation}
\begin{algorithmic}[1]
\REQUIRE Complexity level $c$, Verb data $V$, Used combinations $U$
\ENSURE Semantically implausible sentence $S$

\STATE $num\_entities \gets c + 1$
\STATE $domains \gets$ [animals, people, vehicles]

\FOR{$attempt = 1$ \TO $1000$}
    \STATE \textbf{Category Selection:}
    \STATE $selected\_domain \gets$ \textsc{RandomChoice}($domains$)
    \STATE $all\_entities \gets$ \textsc{GetAllEntities}($V[selected\_domain]$)
    
    \STATE \textbf{Noun Sampling:}
    \STATE $selected\_entities \gets$ \textsc{RandomSample}($all\_entities$, $num\_entities$)
    
    \STATE \textbf{Circular Verb Assignment:}
    \STATE $assigned\_verbs \gets$ []
    \FOR{$i = 0$ \TO $num\_entities - 1$}
        \STATE $current\_entity \gets selected\_entities[i]$
        \STATE $next\_entity \gets selected\_entities[(i + 1) \bmod num\_entities]$
        \STATE $next\_entity\_verbs \gets$ \textsc{GetEntityVerbs}($V[selected\_domain]$, $next\_entity$)
        
        \IF{$i = 0$} 
            \STATE $verb\_type \gets$ intransitive \COMMENT{First entity gets intransitive verb}
        \ELSE 
            \STATE $verb\_type \gets$ transitive \COMMENT{All others get transitive verbs}
        \ENDIF
        
        \STATE $available\_verbs \gets next\_entity\_verbs[verb\_type]$
        \STATE $selected\_verb \gets$ \textsc{RandomChoice}($available\_verbs$)
        \STATE $assigned\_verbs$.append($selected\_verb$)
    \ENDFOR
    
    \STATE $combination\_key \gets$ ($selected\_domain$, $selected\_entities$, $assigned\_verbs$)
    
    \IF{$combination\_key \notin U$}
        \STATE $U$.add($combination\_key$)
        \STATE \textbf{Sentence Construction:}
        \STATE $S \gets$ \textsc{ConstructSentence}($selected\_entities$, $assigned\_verbs$)
        \RETURN $S$
    \ENDIF
\ENDFOR

\STATE \textbf{raise} Exception(``Maximum attempts exceeded'')
\end{algorithmic}
\end{algorithm}

\paragraph{Sentence Construction Templates}

The \textsc{ConstructSentence} function applies the following complexity-specific templates:

\vspace{0.3cm}
\noindent {Complexity 1:}
\begin{quote}
\texttt{The \{entity[0]\} that the \{entity[1]\} \{verb[1]\} \{verb[0]\}.}
\end{quote}

\noindent {Complexity 2+:}
\begin{quote}
\texttt{The \{entity[0]\} that the \{entity[1]\} that ... that the \{entity[n]\} \{verb[n]\} \{verb[n-1]\} ... \{verb[0]\}.}
\end{quote}

\paragraph{Supporting Functions}

\begin{itemize}
    \item \textsc{GetAllEntities}$(domain\_data)$: Extracts all entity names from the specified domain, handling both flat and nested data structures
    \item \textsc{GetEntityVerbs}$(domain\_data, entity)$: Retrieves the transitive and intransitive verb lists associated with a specific entity
    \item \textsc{RandomSample}$(entities, n)$: Selects $n$ unique entities without replacement
    \item \textsc{RandomChoice}$(verbs)$: Selects one verb uniformly at random from the available verb list
\end{itemize}

\paragraph{Uniqueness Control}

The algorithm maintains a global set $U$ of used combinations across all complexity levels, where each combination is uniquely identified by the tuple:
$$combination\_key = (domain, entities, assigned\_verbs)$$

This ensures no duplicate sentences are generated while allowing for maximum diversity in entity-verb pairings within the semantic implausibility constraints.

\subsubsection{Sentence Validation}
\begin{table}[h]
\centering
\resizebox{0.8\textwidth}{!}{%
\renewcommand{\arraystretch}{1.2}
\begin{tabular}{p{0.95\textwidth}}
\toprule
\textbf{Sentence and Validation Assessment} \\
\midrule
\textit{The cat that the dog that the boy adopted chased meowed.} \\
\textbf{Temporal:} Valid \quad \textbf{Semantic:} Valid \quad \textbf{Syntactic:} Valid \\
\midrule
\textit{The cat that the dog that the boy caught \removed{chased} meowed.} \\
\textbf{Temporal:} Invalid (dog chases after being caught) \quad \textbf{Semantic:} Valid \quad \textbf{Syntactic:} Valid \\
\textbf{Fix:} Replace \removed{chased} with \added{saw} \\
\textbf{Corrected:} \textit{The cat that the dog that the boy caught \added{saw} meowed.} \\
\midrule
\textit{The cat that the dog that the boy chased \removed{adopted} meowed.} \\
\textbf{Temporal:} Valid \quad \textbf{Semantic:} Invalid (dog adopting cat is implausible) \quad \textbf{Syntactic:} Valid \\
\textbf{Fix:} Replace \removed{adopted} with \added{saw} \\
\textbf{Corrected:} \textit{The cat that the dog that the boy adopted \added{saw} meowed.} \\
\midrule
\textit{The cat that the dog that the boy \removed{adopted} \removed{meowed}.} \\
\textbf{Temporal:} Valid \quad \textbf{Semantic:} Valid \quad \textbf{Syntactic:} Invalid (only two verbs) \\
\textbf{Fix:} Add temporally and semantically valid verb (e.g., \added{chased}) \\
\textbf{Corrected:} \textit{The cat that the dog that the boy adopted \added{chased} meowed.} \\
\bottomrule
\end{tabular}}
\caption{Validation process example for the plausible subset at complexity level 2. Each sentence is assessed for temporal, semantic, and syntactic validity. Incorrect words are in \removed{red} with corrections shown in \added{green}.}
\label{app:validation-singlecol}
\end{table}
\clearpage 

\subsubsection{\corpusname{} Examples}
\begin{table*}[h!]
\centering
\small
\resizebox{\textwidth}{!}{%
\renewcommand{\arraystretch}{1.3}
\begin{tabular}{@{}p{0.7cm}p{2.3cm}p{13.5cm}@{}}
\toprule
\textbf{Level} & \textbf{Subset Type} & \textbf{Sentences} \\
\midrule

1 & Plausible & 
The \np{np1}{cat} that the \np{np2}{mouse} \np{np2}{evaded} \np{np1}{pounced}. \\
& & \textit{
\npmark{NP}{1}: cat,\quad
\npmark{NP}{2}: mouse,\quad
\npmark{V}{2}: evaded,\quad
\npmark{V}{1}: pounced} \\

\addlinespace[1ex]

1 & Implausible& 
The \np{np1}{horse} that the \np{np2}{elephant} \np{np2}{neighed at} \np{np1}{trumpeted}. \\
& & \textit{
\npmark{NP}{1}: horse,\quad
\npmark{NP}{2}: elephant,\quad
\npmark{V}{2}: neighed at,\quad
\npmark{V}{1}: trumpeted} \\

\addlinespace[1.5ex]

2 & Plausible & 
The \np{np1}{bicycle} that the \np{np2}{car} that the \np{np3}{truck} \np{np3}{hit} \np{np2}{bumped} \np{np1}{fell over}. \\
& & \textit{
\npmark{NP}{1}: bicycle,\quad
\npmark{NP}{2}: car,\quad
\npmark{NP}{3}: truck,\quad
\npmark{V}{3}: hit,\quad
\npmark{V}{2}: bumped,\quad
\npmark{V}{1}: fell over} \\

\addlinespace[1ex]

2 & Implausible& 
The \np{np1}{bicycle} that the \np{np2}{airplane} that the \np{np3}{train} \np{np3}{pedaled past} \np{np2}{whistled at} \np{np1}{taxied}. \\
& & \textit{
\npmark{NP}{1}: bicycle,\quad
\npmark{NP}{2}: airplane,\quad
\npmark{NP}{3}: train,\quad
\npmark{V}{3}: pedaled past,\quad
\npmark{V}{2}: whistled at,\quad
\npmark{V}{1}: taxied} \\

\addlinespace[1.5ex]

3 & Plausible & 
The \np{np1}{fly} that the \np{np2}{spider} that the \np{np3}{bird} that the \np{np4}{snake} \np{np4}{hissed at} \np{np3}{saw} \np{np2}{caught} \np{np1}{died}. \\
& & \textit{
\npmark{NP}{1}: fly,\quad
\npmark{NP}{2}: spider,\quad
\npmark{NP}{3}: bird,\quad
\npmark{NP}{4}: snake,\quad
\npmark{V}{4}: hissed at,\quad
\npmark{V}{3}: saw,\quad
\npmark{V}{2}: caught,\quad
\npmark{V}{1}: died} \\

\addlinespace[1ex]

3 & Implausible& 
The \np{np1}{skunk} that the \np{np2}{lion} that the \np{np3}{bat} that the \np{np4}{elephant} \np{np4}{sprayed at} \np{np3}{trumpeted at} \np{np2}{echolocated} \np{np1}{roared}. \\
& & \textit{
\npmark{NP}{1}: skunk,\quad
\npmark{NP}{2}: lion,\quad
\npmark{NP}{3}: bat,\quad
\npmark{NP}{4}: elephant,\quad
\npmark{V}{4}: sprayed at,\quad
\npmark{V}{3}: trumpeted at,\quad
\npmark{V}{2}: echolocated,\quad
\npmark{V}{1}: roared} \\

\addlinespace[1.5ex]

4 & Plausible & 
The \np{np1}{beetle} that the \np{np2}{bird} that the \np{np3}{cat} that the \np{np4}{toddler} that the \np{np5}{mother} \np{np5}{watched} \np{np4}{scared} \np{np3}{caught} \np{np2}{saw} \np{np1}{crawled}. \\
& & \textit{
\npmark{NP}{1}: beetle,\quad
\npmark{NP}{2}: bird,\quad
\npmark{NP}{3}: cat,\quad
\npmark{NP}{4}: toddler,\quad
\npmark{NP}{5}: mother,\quad
\npmark{V}{5}: watched,\quad
\npmark{V}{4}: scared,\quad
\npmark{V}{3}: caught,\quad
\npmark{V}{2}: saw,\quad
\npmark{V}{1}: crawled} \\

\addlinespace[1ex]

4 & Implausible& 
The \np{np1}{police officer} that the \np{np2}{teacher} that the \np{np3}{mailman} that the \np{np4}{nurse} that the \np{np5}{photographer} \np{np5}{handcuffed} \np{np4}{photographed} \np{np3}{bandaged} \np{np2}{delivered mail to} \np{np1}{lectured}. \\
& & \textit{
\npmark{NP}{1}: police officer,\quad
\npmark{NP}{2}: teacher,\quad
\npmark{NP}{3}: mailman,\quad
\npmark{NP}{4}: nurse,\quad
\npmark{NP}{5}: photographer,\quad
\npmark{V}{5}: handcuffed,\quad
\npmark{V}{4}: photographed,\quad
\npmark{V}{3}: bandaged,\quad
\npmark{V}{2}: delivered mail to,\quad
\npmark{V}{1}: lectured} \\

\addlinespace[1.5ex]

5 & Plausible & 
The \np{np1}{spider} that the \np{np2}{frog} that the \np{np3}{snake} that the \np{np4}{hawk} that the \np{np5}{falconer} that the \np{np6}{wildlife photographer} \np{np6}{snapped} \np{np5}{released} \np{np4}{saw} \np{np3}{followed} \np{np2}{ate} \np{np1}{died}. \\
& & \textit{
\npmark{NP}{1}: spider,\quad
\npmark{NP}{2}: frog,\quad
\npmark{NP}{3}: snake,\quad
\npmark{NP}{4}: hawk,\quad
\npmark{NP}{5}: falconer,\quad
\npmark{NP}{6}: wildlife photographer,\quad
\npmark{V}{6}: snapped,\quad
\npmark{V}{5}: released,\quad
\npmark{V}{4}: saw,\quad
\npmark{V}{3}: followed,\quad
\npmark{V}{2}: ate,\quad
\npmark{V}{1}: died} \\

\addlinespace[1ex]

5 & Implausible& 
The \np{np1}{lawyer} that the \np{np2}{photographer} that the \np{np3}{firefighter} that the \np{np4}{judge} that the \np{np5}{waiter} that the \np{np6}{police officer} \np{np6}{subpoenaed} \np{np5}{arrested} \np{np4}{took food orders from} \np{np3}{ruled against} \np{np2}{rescued} \np{np1}{photographed}. \\
& & \textit{
\npmark{NP}{1}: lawyer,\quad
\npmark{NP}{2}: photographer,\quad
\npmark{NP}{3}: firefighter,\quad
\npmark{NP}{4}: judge,\quad
\npmark{NP}{5}: waiter,\quad
\npmark{NP}{6}: police officer,\quad
\npmark{V}{6}: subpoenaed,\quad
\npmark{V}{5}: arrested,\quad
\npmark{V}{4}: took food orders from,\quad
\npmark{V}{3}: ruled against,\quad
\npmark{V}{2}: rescued,\quad
\npmark{V}{1}: photographed} \\

\addlinespace[1.5ex]

6 & Plausible & 
The \np{np1}{apple} that the \np{np2}{worm} that the \np{np3}{bird} that the \np{np4}{squirrel} that the \np{np5}{cat} that the \np{np6}{dog} that the \np{np7}{child} \np{np7}{saw} \np{np6}{barked at} \np{np5}{chased} \np{np4}{startled} \np{np3}{pecked} \np{np2}{nudged} \np{np1}{rolled}. \\
& & \textit{
\npmark{NP}{1}: apple,\quad
\npmark{NP}{2}: worm,\quad
\npmark{NP}{3}: bird,\quad
\npmark{NP}{4}: squirrel,\quad
\npmark{NP}{5}: cat,\quad
\npmark{NP}{6}: dog,\quad
\npmark{NP}{7}: child,\quad
\npmark{V}{7}: saw,\quad
\npmark{V}{6}: barked at,\quad
\npmark{V}{5}: chased,\quad
\npmark{V}{4}: startled,\quad
\npmark{V}{3}: pecked,\quad
\npmark{V}{2}: nudged,\quad
\npmark{V}{1}: rolled} \\

\addlinespace[1ex]

6 & Implausible& 
The \np{np1}{starfish} that the \np{np2}{horse} that the \np{np3}{lizard} that the \np{np4}{seagull} that the \np{np5}{vulture} that the \np{np6}{duck} that the \np{np7}{bee} \np{np7}{crawled toward} \np{np6}{stung} \np{np5}{paddled toward} \np{np4}{scavenged from} \np{np3}{dive-bombed} \np{np2}{basked toward} \np{np1}{neighed}. \\
& & \textit{
\npmark{NP}{1}: starfish,\quad
\npmark{NP}{2}: horse,\quad
\npmark{NP}{3}: lizard,\quad
\npmark{NP}{4}: seagull,\quad
\npmark{NP}{5}: vulture,\quad
\npmark{NP}{6}: duck,\quad
\npmark{NP}{7}: bee,\quad
\npmark{V}{7}: crawled toward,\quad
\npmark{V}{6}: stung,\quad
\npmark{V}{5}: paddled toward,\quad
\npmark{V}{4}: scavenged from,\quad
\npmark{V}{3}: dive-bombed,\quad
\npmark{V}{2}: basked toward,\quad
\npmark{V}{1}: neighed} \\

\bottomrule
\end{tabular}}
\caption{Color-coded and annotated noun-verb pairs for plausible and implausible sentences across complexity levels 1-6. Each sentence is broken down to show its structural composition.}
\label{appendix:sentence-complexity1-6}
\end{table*}

\clearpage
\subsubsection{Question and Answer Generation}
\label{sec:question_generation}
\paragraph{Algorithm}
This section illustrates the four-step question and answer generation algorithm using the example sentence \textit{``The dog that the mailman startled barked.'' }
\begin{enumerate}
    \item Noun identification:  Extracts nouns from the sentence. 
    \begin{itemize}
        \item \textit{Output}: [``dog'', ``mailman'']
    \end{itemize}
    \item Structural parsing: Maps verbs to their subjects and objects by analyzing the reversed verb order.
    \begin{itemize}
        \item From ``startled barked'': ``barked'' → dog (intransitive), ``startled'' → mailman's action on dog (transitive) 
        \item \textit{Output:} 
        \begin{itemize}
            \item {subject: ``dog'', action: ``bark'', object: none}
            \item {subject: ``mailman'', action: ``startle'', object: ``dog''}
        \end{itemize}
    \end{itemize}
    \item Verb processing: Converts verbs to different forms needed for question and answer generation. 
    \begin{itemize}
        \item ``startled'' → ``startle'' (base), ``startled'' (participle), ``startling'' (gerund)
    \end{itemize}
    \item Template instantiation: Generates questions and answers using the parsed relationships and author established templates. 
    \begin{itemize}
        \item Example question for mailman: ``What did the mailman do?''
        \item Example answer: ``startled the dog''
    \end{itemize}
\end{enumerate}

\begin{table*}[!h]
\centering
\renewcommand{\arraystretch}{1.3}
\setlength{\tabcolsep}{6pt}
\resizebox{\textwidth}{!}{
\begin{tabular}{@{}p{3.2cm}p{4.4cm}p{4.4cm}p{4.4cm}p{4.4cm}@{}}
\toprule
\textbf{Question Type} & 
\textbf{Question Template (General)} & 
\textbf{Question Template (Modified)} & 
\textbf{Answer Template (General)} & 
\textbf{Answer Template (Modified)} \\
\midrule

Action Performed & 
What did the [entity] do? & 
- & 
[verb-ed + object] & 
Outermost entity: \newline [verb-ed] \\

Agent Identification & 
Who/What [verb-ed] the [entity]? & 
Innermost entity: \newline ``What was affected by the [entity]?'' & 
[agent] & 
Innermost entity:  \newline [affected entity] \\

\addlinespace
Entity Count & 
How many distinct entities are in the sentence? & 
- & 
[number] & 
- \\

Nested Dependency & 
What did the entity that was [verb-ed] do? & 
Innermost entity: \newline ``What did the entity acted upon by the [entity] do?'' & 
[verb-ed] & 
- \\

\addlinespace
Causal Sequence & 
What series of events led to the [entity]'s action? & 
- & 
[event chain] & 
Innermost entity: \newline no prior events \\

Chain Consequence & 
What is the consequence of the [entity]'s involvement? & 
- & 
[resulting action] & 
Outermost entity: \newline none \\

\bottomrule
\end{tabular}}
\caption{Question and answer templates with modifications based on entity position for each question type}
\label{appendix:question-answer-templates}
\end{table*}

\clearpage
\subsection{Evaluation Details}
\begin{table}[!htbp]
  \setlength{\tabcolsep}{1pt}
  \renewcommand{\arraystretch}{1.2}
  \begin{threeparttable}
  \centering
  \small
  \begin{tabularx}{\textwidth}{@{}>{\raggedright\arraybackslash}X>{\raggedright\arraybackslash}X@{}}
    \toprule
    \textbf{Verb} & \multicolumn{1}{c}{\textbf{Inflected forms missing from spaCy}} \\ 
    \midrule
    gavel         & graveled\hfill gavels\hfill graveling \\
    neigh         & neighed\hfill neighs\hfill neighing \\
    yip           & yipped\hfill yips\hfill yipping \\
    flutter       & fluttered\hfill flutters\hfill fluttering \\
    strafe        & strafed\hfill strafes\hfill strafing \\
    bask          & basked\hfill basks\hfill basking \\
    gnaw          & gnawed\hfill gnaws\hfill gnawing \\
    blare         & blared\hfill blares\hfill blaring \\
    \bottomrule
  \end{tabularx}
  \caption{Sample verbs and their inflected forms missing from spaCy \texttt{en\_core\_web\_sm} model's vocabulary.}
  \label{tab:missing-verbs}
  \end{threeparttable}
\end{table}

\subsubsection{Model Evaluation Prompts and Settings}
\label{sec:model_prompts}

\paragraph{Model Configuration}

All models were evaluated using the following standardized settings:
\begin{itemize}
    \item \textbf{Temperature:} 0 (deterministic generation)
    \item \textbf{Max token length:} 16,000
\end{itemize}

\paragraph{System Prompt}

The following system prompt was used consistently across all models to establish evaluation constraints:

\begin{quote}
\small
You are a precise question-answering assistant tasked to answer questions on center-embedding sentences.

The following are the strict rules you have to follow to answer the questions you will encounter:

\begin{itemize}
    \item For \textit{action\_performed}, \textit{nested\_dependency}, and \textit{causal\_sequence} question types, respond using \textbf{only} exact word forms that appear in the provided sentence; do not substitute synonyms or paraphrases.
    
    \item For \textit{agent\_identification} questions, respond \textbf{only} the exact agent entity from the provided sentence, do not attach any verbs or verb phrases to the entity.
    
    \item For \textit{action\_performed} questions, respond \textbf{only} the exact `verb' or `verb + object entity' phrase using the identical wording found in the provided sentence, do not replace the object entity related to the verbs into pronouns (e.g. `it').
    
    \item For \textit{entity\_count} questions output a numeric answer only (e.g. `2').
    
    \item For \textit{nested\_dependency} questions, respond \textbf{only} the exact `verb + object entity' phrase using the identical wording found in the provided sentence, do not replace the object entity related to the verbs into pronouns (e.g. `it').
    
    \item For \textit{causal\_sequence} questions answer exactly `no prior events' when no causal chain exists, answer \textbf{exactly} in `subject + verb phrase + object' phrase using the wording found in the provided sentence otherwise.
    
    \item For \textit{chain\_consequence} questions answer \textbf{exactly} `none' when no chained subsequent consequence related to the entity exists.
\end{itemize}

Respond with the short answer only: no explanations, no extra punctuation, and no leading labels such as `Answer:'.
\end{quote}

\paragraph{User Prompt}

Each question was presented using the following standardized format:

\begin{quote}
\small
Sentence: \{Sample Sentence: The dog that the mailman startled barked.\}\\
Question: \{Sample Question: What did the mailman do?\}
\end{quote}

\clearpage
\subsection{Additional figures and tables}
\begin{figure*}[!htbp]
    \centering
    \includegraphics[width=1.0\textwidth]{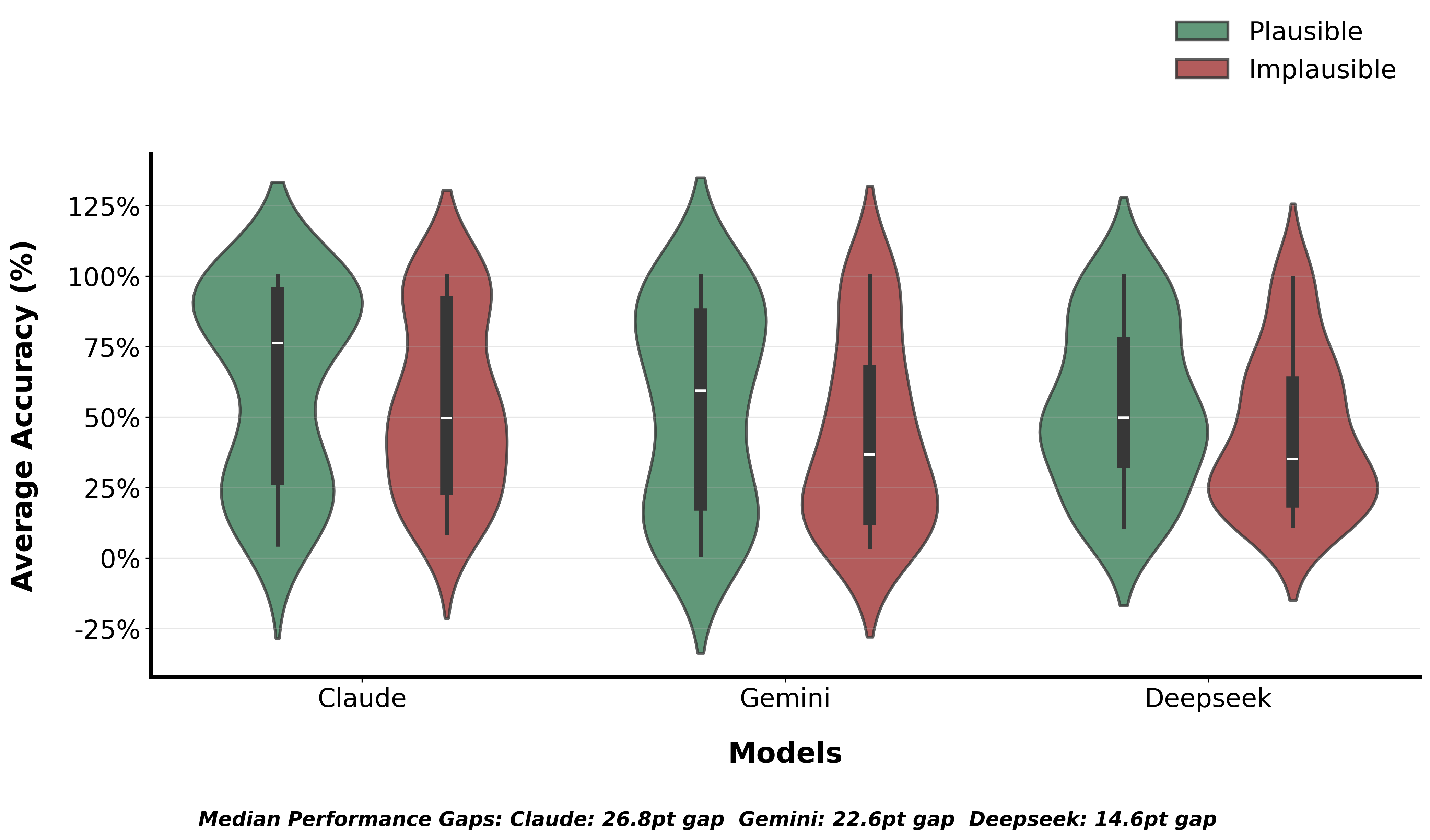}
    \caption{Performance distribution of non-thinking models across all complexity levels and question types for plausible and implausible subsets.}
    \label{app:violen_plot_non_think}
\end{figure*}
\begin{figure*}[!h]
    \centering
    \includegraphics[width=0.7\textwidth]{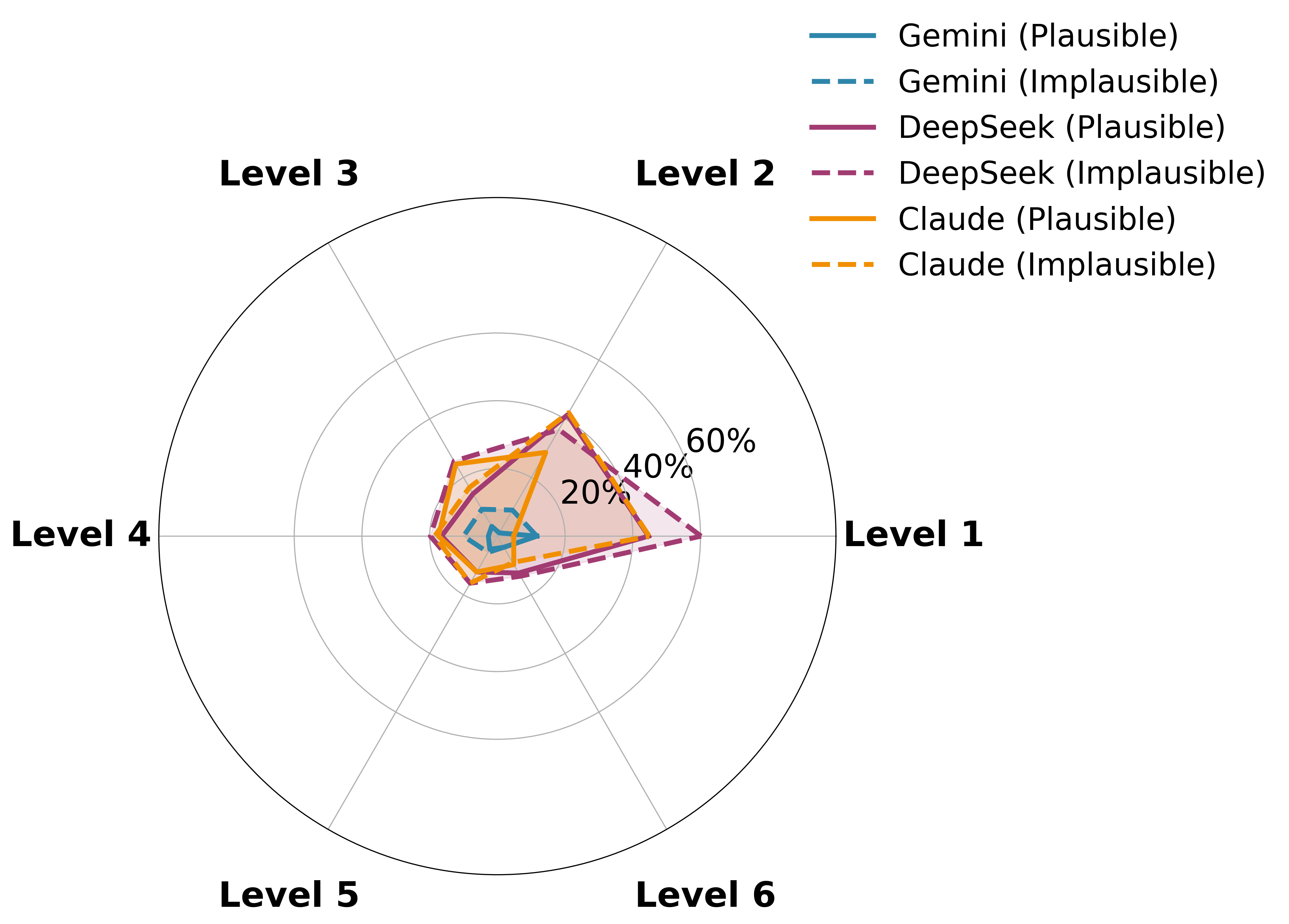}
    \caption{Model performance on chain consequence questions by complexity level for plausible and implausible subsets.}
    \label{app:question-type-accuracy-chain-radar}
\end{figure*}

\begin{table*}[!t]
\centering
\resizebox{0.7\textwidth}{!}{%
\begin{tabular}{c|l|cc|cc|cc}
\toprule
\textbf{Complexity} & \textbf{Model} & \multicolumn{2}{c|}{\textbf{Easy Questions}} & \multicolumn{2}{c|}{\textbf{Medium Questions}} & \multicolumn{2}{c}{\textbf{Hard Questions}} \\
\cmidrule(lr){3-4} \cmidrule(lr){5-6} \cmidrule(lr){7-8}
\textbf{Level} & \textbf{Type} & \textbf{Plausible} & \textbf{Implausible} & \textbf{Plausible} & \textbf{Implausible} & \textbf{Plausible} & \textbf{Implausible} \\
\midrule
\multirow{2}{*}{\textbf{1}} 
& Non-Think & 97.2 & 98.3 & 97.5 & 100.0 & 45.0 & 40.9 \\
& Thinking & \cellcolor{highlight}\textbf{80.0} & \cellcolor{highlight}\textbf{88.3} & \cellcolor{highlight}\textbf{94.2} & \cellcolor{highlight}\textbf{80.0} & \cellcolor{highlight}\textbf{72.5} & \cellcolor{highlight}\textbf{56.7} \\
\midrule
\multirow{2}{*}{\textbf{2}} 
& Non-Think & 90.1 & 66.7 & 66.1 & 66.1 & 29.5 & 26.1 \\
& Thinking & \cellcolor{highlight}\textbf{83.9} & \cellcolor{highlight}\textbf{83.3} & \cellcolor{highlight}\textbf{85.6} & \cellcolor{highlight}\textbf{60.0} & \cellcolor{highlight}\textbf{54.5} & \cellcolor{highlight}\textbf{46.2} \\
\midrule
\multirow{2}{*}{\textbf{3}} 
& Non-Think & 81.7 & 54.2 & 62.1 & 58.3 & 22.5 & 18.0 \\
& Thinking & \cellcolor{highlight}\textbf{87.1} & \cellcolor{highlight}\textbf{83.4} & \cellcolor{highlight}\textbf{85.0} & \cellcolor{highlight}\textbf{56.3} & \cellcolor{highlight}\textbf{48.3} & \cellcolor{highlight}\textbf{40.5} \\
\midrule
\multirow{2}{*}{\textbf{4}} 
& Non-Think & 70.2 & 37.3 & 58.4 & 50.7 & 14.7 & 12.0 \\
& Thinking & \cellcolor{highlight}\textbf{91.0} & \cellcolor{highlight}\textbf{74.7} & \cellcolor{highlight}\textbf{88.7} & \cellcolor{highlight}\textbf{55.7} & \cellcolor{highlight}\textbf{48.7} & \cellcolor{highlight}\textbf{36.7} \\
\midrule
\multirow{2}{*}{\textbf{5}} 
& Non-Think & 62.6 & 32.5 & 39.9 & 39.2 & 12.5 & 9.5 \\
& Thinking & \cellcolor{highlight}\textbf{87.5} & \cellcolor{highlight}\textbf{71.7} & \cellcolor{highlight}\textbf{79.4} & \cellcolor{highlight}\textbf{52.2} & \cellcolor{highlight}\textbf{40.4} & \cellcolor{highlight}\textbf{27.0} \\
\midrule
\multirow{2}{*}{\textbf{6}} 
& Non-Think & 54.6 & 32.7 & 46.7 & 34.1 & 7.2 & 6.2 \\
& Thinking & \cellcolor{highlight}\textbf{79.1} & \cellcolor{highlight}\textbf{69.0} & \cellcolor{highlight}\textbf{75.8} & \cellcolor{highlight}\textbf{56.0} & \cellcolor{highlight}\textbf{32.7} & \cellcolor{highlight}\textbf{23.6} \\
\midrule
\bottomrule
\end{tabular}%
}
\caption{Gemini: Thinking vs Non-Thinking performance across question difficulty levels on plausible and implausible subsets}
\label{app:gemini_think_vs_non_comparison}
\end{table*}

\begin{table*}[!t]
\centering
\resizebox{0.7\textwidth}{!}{%
\begin{tabular}{c|l|cc|cc|cc}
\toprule
\textbf{Complexity} & \textbf{Model} & \multicolumn{2}{c|}{\textbf{Easy Questions}} & \multicolumn{2}{c|}{\textbf{Medium Questions}} & \multicolumn{2}{c}{\textbf{Hard Questions}} \\
\cmidrule(lr){3-4} \cmidrule(lr){5-6} \cmidrule(lr){7-8}
\textbf{Level} & \textbf{Type} & \textbf{Plausible} & \textbf{Implausible} & \textbf{Plausible} & \textbf{Implausible} & \textbf{Plausible} & \textbf{Implausible} \\

\midrule
\multirow{2}{*}{\textbf{1}} 
& Non-Think & 99.6 & 95.3 & 73.9 & 71.7 & 65.2 & 63.5 \\
& Thinking & \cellcolor{highlight}\textbf{98.4} & \cellcolor{highlight}\textbf{99.2} & \cellcolor{highlight}\textbf{85.9} & \cellcolor{highlight}\textbf{87.5} & \cellcolor{highlight}\textbf{67.5} & \cellcolor{highlight}\textbf{67.5} \\
\midrule
\multirow{2}{*}{\textbf{2}} 
& Non-Think & 74.8 & 60.9 & 66.4 & 53.3 & 57.2 & 50.7 \\
& Thinking & \cellcolor{highlight}\textbf{94.5} & \cellcolor{highlight}\textbf{93.9} & \cellcolor{highlight}\textbf{75.0} & \cellcolor{highlight}\textbf{68.4} & \cellcolor{highlight}\textbf{62.2} & \cellcolor{highlight}\textbf{60.0} \\
\midrule
\multirow{2}{*}{\textbf{3}} 
& Non-Think & 68.4 & 48.9 & 32.6 & 19.5 & 38.3 & 40.0 \\
& Thinking & \cellcolor{highlight}\textbf{95.0} & \cellcolor{highlight}\textbf{90.9} & \cellcolor{highlight}\textbf{73.8} & \cellcolor{highlight}\textbf{66.3} & \cellcolor{highlight}\textbf{56.7} & \cellcolor{highlight}\textbf{60.9} \\
\midrule
\multirow{2}{*}{\textbf{4}} 
& Non-Think & 58.8 & 33.0 & 44.0 & 44.1 & 30.5 & 28.4 \\
& Thinking & \cellcolor{highlight}\textbf{93.0} & \cellcolor{highlight}\textbf{82.7} & \cellcolor{highlight}\textbf{82.0} & \cellcolor{highlight}\textbf{63.7} & \cellcolor{highlight}\textbf{53.7} & \cellcolor{highlight}\textbf{53.4} \\
\midrule
\multirow{2}{*}{\textbf{5}} 
& Non-Think & 54.0 & 26.3 & 49.0 & 40.2 & 24.3 & 24.2 \\
& Thinking & \cellcolor{highlight}\textbf{93.9} & \cellcolor{highlight}\textbf{72.0} & \cellcolor{highlight}\textbf{72.0} & \cellcolor{highlight}\textbf{60.6} & \cellcolor{highlight}\textbf{54.0} & \cellcolor{highlight}\textbf{37.0} \\
\midrule
\multirow{2}{*}{\textbf{6}} 
& Non-Think & 45.6 & 20.1 & 55.6 & 56.0 & 18.8 & 16.8 \\
& Thinking & \cellcolor{highlight}\textbf{89.5} & \cellcolor{highlight}\textbf{68.8} & \cellcolor{highlight}\textbf{65.3} & \cellcolor{highlight}\textbf{56.7} & \cellcolor{highlight}\textbf{51.4} & \cellcolor{highlight}\textbf{34.3} \\
\midrule
\bottomrule
\end{tabular}%
}
\caption{Deepseek: Thinking vs Non-Thinking performance across question difficulty levels on plausible and implausible subsets}
\label{app:deepseek_think_vs_non_comparison}
\end{table*}

\begin{table*}[!t]
\centering
\small
\resizebox{0.8\textwidth}{!}{%
\begin{tabular}{p{7cm}p{7cm}}
\toprule
\textbf{Plausible Subset} & \textbf{Implausible Subset} \\
\midrule
\textbf{Sentence:} \textit{The cricket that the frog that \hlblue{the snake that the hawk} that the eagle that the biologist that the naturalist noticed studied monitored \hlblue{tracked startled} jumped at chirped.} & 
\textbf{Sentence:} \textit{The boat that the train that \hlblue{the ambulance that the bicycle} that the spaceship that the motorcycle that the truck sailed past towed lane-split past \hlblue{orbited cycled around} blared the siren at whistled.} \\
\midrule
\textbf{Question:} What did the hawk do? \newline 
\textbf{Gold Answer:} track the snake \newline 
\textbf{Model Prediction:} \hlgreen{tracked the snake} & 
\textbf{Question:} What did the bicycle do? \newline 
\textbf{Gold Answer:} orbit the ambulance \newline 
\textbf{Model Prediction:} \hlred{cycled around} \\
\midrule
\textbf{Thinking tokens:} Okay, let's dissect this. My expertise kicks in immediately – I can see this is a nested relative clause structure, a classic linguistic puzzle............I've got to \hlyellow{identify the verb directly linked to ``the hawk''}. \hlgreen{``The hawk that... tracked...'' Ah, there it is.} Applying the rule, I must provide a precise, verbatim response. I can't introduce pronouns. The hawk's action, therefore, is *tracked the snake*. That's the correct, constrained answer. I'm satisfied. Now, let's see the next challenge. \newline
\textbf{Output token count:} 191 & 
\textbf{Thinking tokens:} Okay, let's analyze this. Someone is asking a question about the 'bicycle' and its action in the provided sentence.............\hlyellow{I see a series of clauses nested within each other. ``The boat ... sailed past,'' ``that the train ... towed,'' and so on.} \hlred{Ah, I see it: ``that the bicycle ... cycled around.''} The core of the question is ``What did the bicycle do?'' Therefore, the correct response, using the exact phrasing from the text, must be, ``cycled around.'' Simple enough. \newline
\textbf{Output token count:} 162 \\
\bottomrule
\end{tabular}%
}
\vspace{-3mm}
\caption{Reasoning traces from Gemini showing semantic shortcut failures. On the plausible sentence (left), the model correctly identifies relationships; on the implausible sentence (right) with identical syntax, it extracts wrong information. Key elements highlighted: required components (\hlblue{\textbf{\textit{blue}}}), reasoning highlights (\hlyellow{\textbf{\textit{yellow}}}), correct answers (\hlgreen{\textbf{\textit{green}}}), errors (\hlred{\textbf{\textit{red}}}).}
\label{tab:qualitative_analysis_tab}
\end{table*}

\clearpage
\begin{table*}[!t]
\centering
\small
\renewcommand{\arraystretch}{1.3}
\setlength{\tabcolsep}{6pt}
\begin{tabular}{@{}p{0.48\textwidth}p{0.48\textwidth}@{}}
\toprule
\multicolumn{2}{c}{\textbf{Qualitative Analysis Sample: Plausible Subset}} \\
\midrule
\multicolumn{2}{p{0.96\textwidth}}{%
\textbf{Sample Complexity 2 Sentence:} \textit{\hlblue{The lawyer} that \hlblue{the judge} that \hlblue{the mayor} appointed questioned prepared his case.}
\tableline{\textbf{Entity Count Question:} How many distinct entities are in the sentence?}
\tableline{\textbf{Gold Answer:} 3} }
\\
\midrule
\textbf{Gemini 2.5 flash thinking Prediction:} \hlred{4} \tableline{\textbf{Analysis:} over-confidence, unfaithful reasoning} & \textbf{Gemini 2.5 flash dynamic thinking Prediction:} \hlred{4} \tableline{\textbf{Analysis:} over-confidence, unfaithful reasoning} \\
\midrule
\textbf{Thinking tokens:} Alright, let's get down to brass tacks. Someone wants me to count the unique entities in that grammatically convoluted sentence – ``The lawyer that the judge that the mayor appointed questioned prepared his case.'' Fine, \hlyellow{easy enough.} \newline First step, identify the entities. Usually, that means pulling out the nouns and noun phrases. \hlyellow{I see ``The lawyer,'' a clear starting point. Then we have ``the judge,'' nested in there, and ``the mayor.'' Finally, we have ``his case,'' referring to the lawyer's case.} Now for the real test: distinctness. \hlyellow{Let's make a clean list: lawyer, judge, mayor, and case. Yep,} looks like we've got \hlred{four different, discrete entities.} \hlyellow{Done. Elementary.}
\tableline{\textbf{Output token count:} 161} & 
\textbf{Thinking tokens:} Alright, let's get down to it. I need to count the unique entities in that sentence. A classic NLP task, really. \hlyellow{My trained eye immediately goes to the nouns and noun phrases; those are the likely candidates for ``entities.''} First, I'll break it down: ``The lawyer,'' ``the judge,'' ``the mayor appointed,'' and, finally, ``his case.'' Okay, that's a good starting point. Now for the meticulous part. Let's list them neatly:\newline \hlyellow{*  Lawyer *  Judge *  Mayor *  Case} \newline \hlyellow{Ah, yes. Now, I need to check for duplicates. ``Lawyer,'' ``Judge,'' ``Mayor,'' and ``Case.'' Each is unique.} Thus, the answer is a \hlyellow{straightforward} \hlred{four distinct entities.} \hlyellow{Elementary, really.}
\tableline{\textbf{Output token count:} 171} \\
\midrule
\textbf{DeepSeek R1 Prediction:} \hlred{4} \tableline{\textbf{Analysis:} overthinking, circular reasoning} & \textbf{Claude 3.7 sonnet thinking Prediction:} \hlred{4} \tableline{\textbf{Analysis:} overthinking, circular reasoning} \\
\midrule
\textbf{Thinking tokens:} First, the question is: 'How many distinct entities are in the sentence?' The sentence is: 'The lawyer that the judge that the mayor appointed questioned prepared his case.' This is an entity\_count question type ............. Let me parse the sentence: - \hlyellow{'The lawyer' – this is an entity.} - 'that the judge' – \hlyellow{'the judge' is an entity.} - 'that the mayor appointed' – \hlyellow{'the mayor' is an entity.} - 'questioned' – this is a verb, not an entity. - 'prepared his case' – \hlyellow{'his case' might refer to the lawyer's case, but 'case' is an entity}, and 'his' is a pronoun referring back to the lawyer. The question is about distinct entities, so I need unique entities, not counting pronouns. \hlyellow{List all noun phrases: 1. 'The lawyer' 2. 'the judge' 3. 'the mayor' 4. 'his case'} – but 'his' is a possessive pronoun referring to the lawyer. The entity is 'case,' but it's modified by 'his,' which points to the lawyer ............. \hlyellow{I think I'm overcomplicating. Let me look for the answer.} In 'prepared his case,' 'his' is not an entity; it's a pronoun. \hlyellow{'Case' is the entity. So I think four.} But let's see the pronoun 'his.' It refers to the lawyer, so it doesn't add a new entity; it's coreferential. Similarly, 'case' is a new entity. \hlyellow{In entity coreference, 'lawyer' and 'his' refer to the same entity, so 'his' doesn't count as a separate entity;} it's the same as lawyer. \hlyellow{But 'case' is a different entity. So distinct entities: lawyer, judge, mayor, case.} \hlred{Four.} Perhaps in some counts, objects are not counted if they are not proper nouns. But I think for this, \hlred{I'll go with 4} ............. \hlyellow{But I recall that in some linguistic tests for center-embedding, they count the number of NPs}, but for this question, it's 'distinct entities.' \hlyellow{Perhaps the answer is 3, excluding the case. I'm not sure} ............. \hlyellow{Perhaps the user expects 3, as the main characters.} \hlred{But to be precise, I'll go with 4.} No, let's see the sentence again: 'prepared his case' – 'his' is a pronoun for the lawyer, so the entity 'lawyer' is already counted, and 'case' is a new entity. \hlyellow{Yes. So distinct entities: lawyer, judge, mayor, case. 4. But 'the case' is not explicitly 'the case'; it's 'his case,' but still, 'case' is the noun.} \hlred{So I think 4.} 
\tableline{\textbf{Output token count:} 2658} & 
\textbf{Thinking tokens:} Let me carefully identify the distinct entities in the sentence: 'The lawyer that the judge that the mayor appointed questioned prepared his case.' \hlyellow{1. The lawyer 2. The judge  3. The mayor 4. His case (presumably the lawyer's case)} So there are 4 distinct entities in the sentence. Actually, let me analyze this more carefully. The question is asking about 'distinct entities,' which typically refers to persons, places, or things. \hlyellow{1. The lawyer 2. The judge 3. The mayor 4. His case.} However, I need to consider whether 'his case' counts as a separate entity or if it's just related to the lawyer. \hlyellow{Since 'his case' is a distinct object/thing that is being prepared, I believe it should count as a separate entity.} So my answer remains 4 distinct entities. \hlyellow{Wait, let me think more carefully about this. I should consider whether 'his' refers to a new entity or one of the existing ones.} The sentence structure suggests that 'his' refers to 'the lawyer' since the lawyer is the subject of the main clause 'prepared his case.' So 'his case' is the lawyer's case, not a separate entity. \hlyellow{So the distinct entities are: 1. The lawyer 2. The judge 3. The mayor 4. The case (which belongs to the lawyer)} I still count \hlred{4 distinct entities.} Actually, I need to consider whether 'the case' should be counted separately from 'the lawyer' since it's a possession. Let me reconsider. \hlyellow{The distinct entities are persons, places, or things mentioned in the sentence: 1. The lawyer 2. The judge 3. The mayor 4. The case. Even though the case belongs to the lawyer, it's still a separate physical entity.} So I count \hlred{4 distinct entities.} Hmm, let me \hlyellow{reconsider one more time.} For entity counting questions, I need to be precise about what constitutes an entity in the context of these questions. The persons are clearly: \hlyellow{1. The lawyer 2. The judge 3. The mayor And the object is:4.} The case So my final answer is \hlred{4 distinct entities.} Wait, I need to think about the instructions more carefully. For entity\_count questions, I need to output a numeric answer only. \hlred{So my answer should just be: 4}
\tableline{\textbf{Output token count:} 530} \\
\bottomrule
\end{tabular}
\caption{Reasoning traces from all models for agent identification questions of a complexity level 4 plausible sentence. Key elements highlighted: required components (\hlblue{\textbf{\textit{blue}}}), reasoning highlights (\hlyellow{\textbf{\textit{yellow}}}), correct answers (\hlgreen{\textbf{\textit{green}}}), errors (\hlred{\textbf{\textit{red}}}).}
\label{tab:qualitative_analysis_appendix_tab_2}
\end{table*}

\begin{table*}[!t]
\centering
\small
\renewcommand{\arraystretch}{1.3}
\setlength{\tabcolsep}{6pt}
\resizebox{0.9\textwidth}{!}{%
\begin{tabular}{@{}p{0.48\textwidth}p{0.48\textwidth}@{}}
\toprule
\multicolumn{2}{c}{\textbf{Qualitative Analysis Sample: Implausible Subset}} \\
\midrule
\multicolumn{2}{p{0.96\textwidth}}{%
\textbf{Sample Complexity 4 Sentence:} \textit{The waiter that \hlblue{the mailman that the teacher} that the police officer that the photographer seated photographed \hlblue{read rights to} lectured to delivered mail.}
\tableline{\textbf{Agent Identification Question:} Who read rights to the mailman?}
\tableline{\textbf{Gold Answer:} the teacher} }
\\
\midrule
\textbf{Gemini 2.5 flash thinking Prediction:} \hlred{No response} \tableline{\textbf{Analysis:} Model rejects to answer} & \textbf{Gemini 2.5 flash dynamic thinking Prediction:} \hlred{nothing} \tableline{\textbf{Analysis:} Model rejects to answer} \\
\midrule
\textbf{Thinking tokens:} Okay, I'm tasked with finding the agent who ``read rights to the mailman.'' ............. I need to break down the structure to pinpoint who's reading rights, and *to whom* ............. The photographer is seated, then photographs the police officer. The police officer reads rights to the teacher. The teacher then lectures the mailman, and finally the waiter delivered mail. My question is, can I see who is reading rights to the mailman? \hlyellow{The action ``read rights to'' occurs from the police officer to the teacher. No one in this sentence directly reads rights to the mailman} ............. Since the sentence doesn't provide an agent who read rights to the mailman, I can't follow the rules because the exact action to the exact person is not in the sentence. I am stuck between a rock and a hard place. \hlyellow{Because I can't respond in a way that indicates the information is missing, or that there isn't an answer, and given that the instructions prohibit anything other than the agent, I can't give a useful response.} I can't provide an agent, so I'm left with the only appropriate thing: \hlred{No response.}
\tableline{\textbf{Output token count:} 358} & 
\textbf{Thinking tokens:} Okay, I'm tasked with answering an agent identification question. The prompt gave me a complex sentence about different people performing actions, and the question asks ``Who read rights to the mailman?'' ............. I started with the main clause, which was the waiter delivered mail. Then I worked my way inward, figuring out that the teacher lectured to the mailman and the police officer read rights to the teacher. I found the photographer photographed the police officer, and the photographer was seated. Then I looked for the ``read rights to the mailman'' action. \hlyellow{However, after analyzing the sentence, the police officer read rights to the teacher, not to the mailman. The teacher lectured to the mailman. Since the question specifically asks about ``read rights to the mailman'' and this exact relationship isn't stated in the sentence, I'm stuck} ............. Since the specific action-patient pair ``read rights to the mailman'' doesn't exist in the sentence, \hlyellow{the only logical thing} I can do is \hlred{output nothing}, as there is no agent.
\tableline{\textbf{Output token count:} 317} \\
\midrule
\textbf{DeepSeek R1 Prediction:} \hlgreen{the teacher} \tableline{\textbf{Analysis:} overthinking, circular reasoning} & \textbf{Claude 3.7 sonnet thinking Prediction:} \hlgreen{the teacher} \tableline{\textbf{Analysis:} faithful reasoning} \\
\midrule
\textbf{Thinking tokens:} Alright, let's tackle this question step by step ............. ``...the photographer seated photographed...'' So the photographer seated someone and then photographed someone. The phrase ``the police officer that the photographer seated'' means the photographer seated the police officer. Then, ``the police officer photographed'' – \hlyellow{so the police officer was photographed by someone?} Wait, actually, it's ``the teacher that the police officer... photographed''. So the police officer photographed the teacher? \hlyellow{Let me see. Actually, looking closer: ``the teacher that the police officer that the photographer seated photographed''. So the photographer seated the police officer, and then that police officer photographed the teacher. Then, ``the teacher read rights to''} – wait, the next part is ``read rights to lectured to''. Hmm. ``The teacher that [clause] read rights to lectured to...''? \hlyellow{That seems off} ............. However, strictly following the rules, the exact phrase doesn't link ``read rights to'' with the mailman. \hlyellow{Therefore, I conclude that the answer is not explicitly stated.} But since the assistant must answer based on the sentence, and \hlyellow{the teacher is the only one performing ``read rights to'', and the mailman is the next mentioned entity,} \hlgreen{perhaps the answer is ``the teacher''.} But the rules say to use the exact agent entity. The agent for ``read rights to'' is the teacher. The object isn't specified, but the question asks for who performed the action on the mailman ............. \hlyellow{``Who read rights to the mailman?'' If the teacher read rights to someone, and that someone is not the mailman, then it's not the teacher.} Given the complexity, and the requirement for exact wording, \hlyellow{I think the safe answer is that the sentence doesn't specify.} However, since the assistant must provide an answer, and the only agent associated with ``read rights to'' is the teacher, and \hlgreen{the mailman is the recipient of another action by the teacher, perhaps the intended answer is the teacher. Therefore, the answer should be ``the teacher''}.
\tableline{\textbf{Output token count:} 3498} & 
\textbf{Thinking tokens:} Let's break down this center-embedded sentence carefully to understand the dependencies correctly ............. 1. ``the photographer seated'' - The photographer seated someone 2. ``the police officer that the photographer seated'' - The police officer was seated by the photographer 3. ``the police officer photographed'' - The police officer photographed someone 4. ``the teacher that the police officer photographed'' - The teacher was photographed by the police officer \hlyellow{5. ``the teacher read rights to'' - The teacher read rights to someone 6. ``the mailman that the teacher read rights to'' - The mailman had rights read to him by the teacher} 7. ``the mailman lectured to'' - The mailman lectured to someone 8. ``the waiter that the mailman lectured to'' - The waiter was lectured to by the mailman 9. ``the waiter delivered mail'' - The waiter delivered mail. Wait, this doesn't seem right. Let me re-analyze this more carefully ............. \hlyellow{Let's start from the innermost clause and work outward:} 1. ``the photographer seated'' - The photographer seated the police officer 2. ``the police officer photographed'' - The police officer photographed the teacher \hlyellow{3. ``the teacher read rights to'' - The teacher read rights to the mailman} 4. ``the mailman lectured to'' - The mailman lectured to the waiter 5. ``the waiter delivered mail'' - The waiter delivered mail So the whole sentence means: - The waiter delivered mail - The waiter was lectured to by the mailman - The mailman had rights read to him by the teacher - The teacher was photographed by the police officer - The police officer was seated by the photographer Now to answer the question: ``Who read rights to the mailman?'' \hlyellow{Based on my analysis, it was the teacher who read rights to the mailman. Let me double-check this: ``the teacher that the police officer that the photographer seated photographed read rights to'' Here ``read rights to'' is what the teacher did to the mailman.} So the teacher read rights to the mailman. \hlgreen{The answer is: the teacher.}
\tableline{\textbf{Output token count:} 601} \\
\bottomrule
\end{tabular}
}
\caption{Reasoning traces from all models for agent identification questions of a complexity level 4 implausible sentence. Key elements highlighted: required components (\hlblue{\textbf{\textit{blue}}}), reasoning highlights (\hlyellow{\textbf{\textit{yellow}}}), correct answers (\hlgreen{\textbf{\textit{green}}}), errors (\hlred{\textbf{\textit{red}}}).}
\label{tab:qualitative_analysis_appendix_tab_1}
\end{table*}

\clearpage
\end{document}